\documentclass[10pt,twocolumn,letterpaper]{article}

\usepackage[pagenumbers]{cvpr} %

\usepackage{graphicx}
\usepackage{amsmath}
\usepackage{amssymb}
\usepackage{booktabs}

\usepackage{accents}
\usepackage{url}
\usepackage{makecell}
\usepackage{amsfonts}       %
\usepackage{nicefrac}       %
\usepackage{microtype}      %
\usepackage{color}
\usepackage{tabularx}
\usepackage[linesnumbered,ruled]{algorithm2e}

\SetCommentSty{mycommfont}
\usepackage{times}
\usepackage{epsfig}
\usepackage{graphicx}
\usepackage{amsmath}
\usepackage{amssymb}
\usepackage{enumitem}
\usepackage{graphicx}
\usepackage{xspace}
\usepackage{balance}
\usepackage{multirow}
\usepackage{mathtools}
\usepackage{rotating}
\usepackage{tabulary}
\usepackage{array}
\usepackage{subcaption}
\usepackage[export]{adjustbox}
\usepackage{ctable}
\usepackage{physics}
\usepackage[title]{appendix}
\usepackage{xcolor}
\usepackage{soul}
\usepackage{newclude}

\usepackage[pagebackref,breaklinks,colorlinks]{hyperref}

\usepackage[capitalize]{cleveref}
\crefname{section}{Sec.}{Secs.}
\Crefname{section}{Section}{Sections}
\Crefname{table}{Table}{Tables}
\crefname{table}{Tab.}{Tabs.}

\def\cvprPaperID{4872} %
\def\confName{CVPR}
\def\confYear{2023}

\begin{document}

\title{Asymmetric Co-teaching with Multi-view Consensus \\ for Noisy Label Learning}
\author{
\parbox{1.0\linewidth}{\centering $\quad$ Fengbei Liu\textsuperscript{\rm 1} $\quad$  Yuanhong Chen\textsuperscript{\rm 1} $\quad$ Chong Wang \textsuperscript{\rm 1} $\quad$   Yu Tian\textsuperscript{\rm 2}  $\quad$ Gustavo Carneiro\textsuperscript{\rm 3} $\newline$   
\textsuperscript{\rm 1} Australian Institute for Machine Learning, University of Adelaide \\
\textsuperscript{\rm 2} Harvard Medical School, Harvard University}  \\
\textsuperscript{\rm 3} CVSSP, University of Surrey
}
\maketitle

\begin{abstract}
Learning with noisy-labels has become an important research topic in computer vision where state-of-the-art (SOTA) methods explore: 1) prediction disagreement with co-teaching strategy that updates two models when they disagree on the prediction of training samples; and 2) sample selection to divide the training set into clean and noisy sets based on small training loss. However, the quick convergence of co-teaching models to select the same clean subsets combined with relatively fast overfitting of noisy labels may induce the wrong selection of noisy label samples as clean, leading to an inevitable confirmation bias that damages accuracy.
In this paper, we introduce our noisy-label learning approach, called Asymmetric
Co-teaching (AsyCo), which introduces novel prediction disagreement that produces more consistent divergent results of the co-teaching models, and a new sample selection approach that does not require small-loss assumption to enable a better robustness to confirmation bias than previous methods.
More specifically, the new prediction disagreement is achieved with the use of different training strategies, where one model is trained with multi-class learning and the other with multi-label learning. Also, the new sample selection is based on multi-view consensus, which uses the label views from training labels and model predictions to divide the training set into clean and noisy for training the multi-class model and to re-label the training samples with multiple top-ranked labels for training the multi-label model.
Extensive experiments on synthetic and real-world noisy-label datasets show that AsyCo improves over current SOTA methods.
\end{abstract}

\vspace{-0.1cm}
\section{Introduction}
\label{sec:intro}

\begin{figure}
    \centering
    \includegraphics[width=\linewidth]{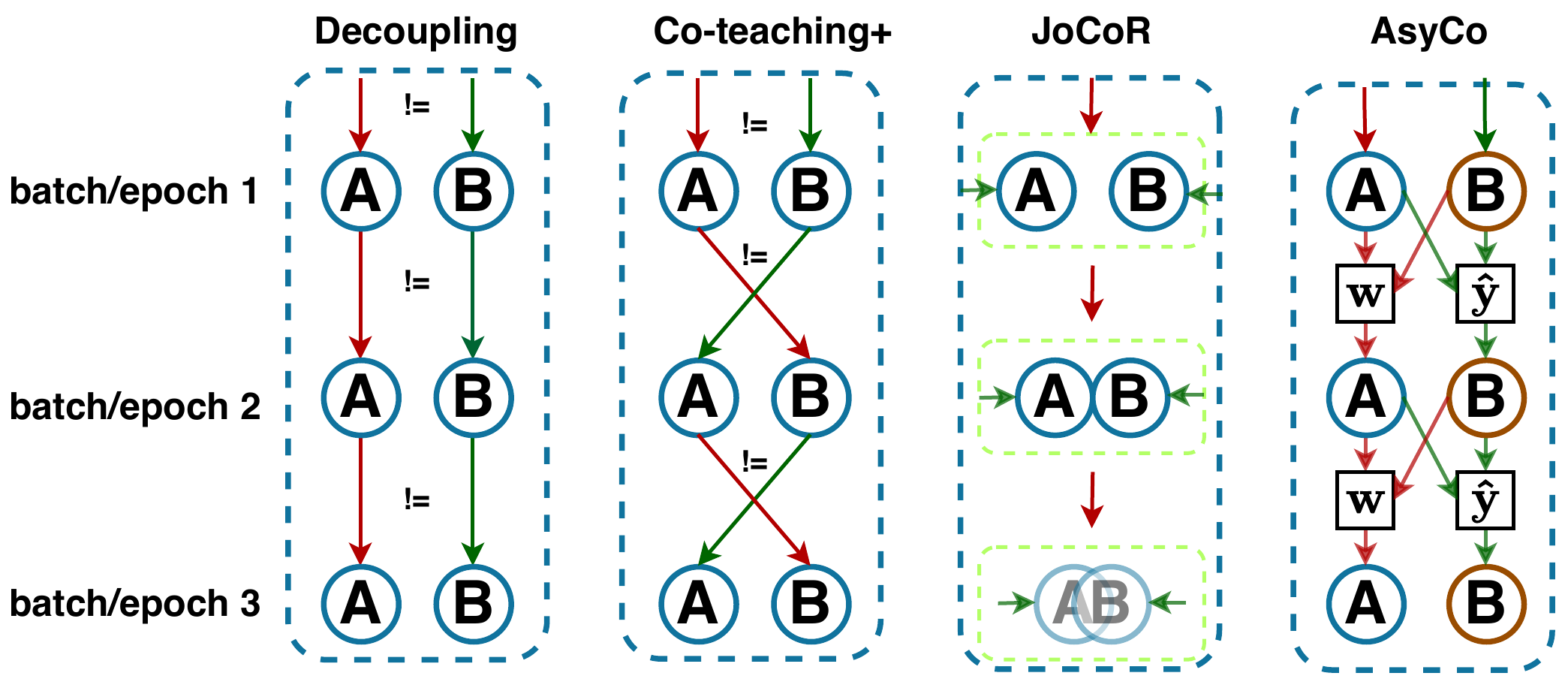}
    \caption{Comparison of methods Decoupling~\cite{malach2017decoupling}, Co-teaching+~\cite{yu2019does},  JoCoR~\cite{wei2020combating}, and our AsyCo. 
    AsyCo co-teaches the multi-class model A and the multi-label model B with different training strategies (denoted by the different colours of A\&B). The training samples for A and B, represented by the green and red arrows, are formed by our proposed multi-view consensus that uses label views from the training set and model predictions to estimate the  variables $\mathbf{w}$ and $\hat{\mathbf{y}}$, which selects clean/noisy samples for training A and iteratively re-labels samples for training B, respectively.}
    \label{fig:Fig1}
\end{figure}

Deep neural network (DNN) has achieved remarkable success in many fields, including computer vision~\cite{krizhevsky2017imagenet,he2015deep}, natural language processing (NLP)~\cite{devlin2018bert,young2018recent} and medical image analysis~\cite{litjens2017survey,wang2017chestx}. However, the methods from those fields often require massive amount of high-quality annotated data for supervised training~\cite{deng2009imagenet}, which is challenging and expensive to acquire. 
To alleviate such problem, some datasets have been annotated via crowdsourcing~\cite{xiao2015learning}, from search engines~\cite{song2019selfie}, or with NLP  from radiology reports~\cite{wang2017chestx}. 
Although these cheaper annotation processes enable the construction of large-scale datasets, they inevitably introduce noisy labels for model training, resulting in DNN model performance degradation. 
Therefore, novel learning algorithms are required to robustly train DNN models when training sets containing noisy labels.

Previous methods tackle noisy-label learning from different perspectives. 
For example, some approaches focus on \textit{prediction disagreement }~\cite{yu2019does,wei2020combating,malach2017decoupling}, which rely on jointly training two models to update their parameters when they disagree on the predictions of the same training samples. 
These two models generally use the same training strategy, so even though they are trained using samples with divergent predictions, both models will quickly converge to select similar clean samples during training, which neutralises the effectiveness of prediction disagreement.
Other noisy-label learning methods are based on \textit{sample selection}~\cite{li2020dividemix,han2018co,arazo2019unsupervised} to find clean and noisy-label samples that are treated differently in the training process. 
Sample-selection approaches usually assume that samples with small training losses are associated with clean labels, which is an assumption verified only at early training stages~\cite{liu2020early,zhang2021understanding}. 
However, such assumption is unwarranted in later training stages because DNN models can overfit any type of noisy label after a certain number of epochs, essentially reducing the training loss for all training samples.
State-of-the-art (SOTA) noisy-label learning approaches~\cite{li2020dividemix} have been designed to depend on both prediction disagreement and sample selection methods to achieve better performance than either method alone. 
Nevertheless, these SOTA methods are still affected by the fast convergence of both models and label noise overfitting, which raises the following questions: 
1) Are there more effective ways to maximise the prediction disagreement between both models, so they consistently produce divergent results during the training procedure?
2) Is there a sample selection approach that can better integrate prediction disagreements than the small loss strategy?

Motivated by traditional multi-view learning~\cite{blum1998combining,sindhwani2005co} and multi-label learning~\cite{shi2020multi}, we propose a new noisy-label learning method that aims to answer the two questions above.
Our method, named \textbf{Asymmetric Co-teaching (AsyCo)} and depicted in Fig.~\ref{fig:Fig1}, is based on two models trained with different learning strategies to maximise their prediction disagreement. 
One model, the \textbf{classification net}, is trained with conventional multi-class learning by minimising a cross entropy loss and provide single-class prediction, and the other, the \textbf{reference net}, is trained with a binary cross entropy loss to enable multi-label learning that is used to estimate the top-ranked labels that represent the potentially clean candidate labels for each training sample.
The original training labels and the predictions by the training and reference nets enable the formation of three label views for each training sample, allowing us to formulate the \textbf{multi-view consensus} that is tightly integrated with the prediction disagreement to select clean and noisy samples for training the multi-class model and to iteratively re-label samples with multiple top-ranked labels for training the multi-label model.
In summary, our main contributions are:
\begin{itemize}
    \item The new noisy-label co-teaching method \textbf{AsyCo} designed to maximise the prediction disagreement  between the training of a multi-class and a multi-label model; and
    \item The novel \textbf{multi-view consensus} that uses the disagreements between training labels and model predictions to select clean and noisy samples for training the multi-class model and to iteratively re-label samples with multiple top-ranked labels for training the multi-label model.
\end{itemize}
We conduct extensive experiments on both synthetic and real-world noisy datasets that show that AsyCo provides substantial improvements over previous state-of-the-art (SOTA) methods.

\section{Related Work}
\label{sec:related}

\textbf{Prediction disagreement} approaches seek to maximise model performance by exploring the prediction disagreements between models trained from the same training set. 
In general, these methods~\cite{malach2017decoupling,yu2019does,wei2020combating, jiang2018mentornet} train two models using samples that have different predictions from both models to mitigate the problem of confirmation bias (i.e., a mistake being reinforced by further training from the same mistake) that particularly affects single-model training.  
Furthermore, the cross teaching of two models can help escape local minima.
Most of the prediction-disagreement methods also rely on sample-selection techniques, as we explain below, but in general, they 
use the same training strategy to train two models, which limits the ability of these approaches to maximise the divergence between the models.

\textbf{Sample selection} approaches aim to automatically classify training samples into clean or noisy and treat them differently during the training process.
Previous papers~\cite{liu2020early,zhang2021understanding} have shown that when training with noisy label, DNN fits the samples with clean labels first and gradually overfits the samples with noisy labels later.
Such training loss characterisation allowed researchers to assume that samples with clean labels have small losses, particularly at early training stages -- this is known as the \textit{small-loss assumption}.
For examples, M-correction~\cite{arazo2019unsupervised} automatically selects clean samples by modelling the training loss distribution with a Beta Mixture model (BMM).
Sample selection has been combined with prediction disagreement in several works, such as Co-teaching~\cite{han2018co} and Co-teaching+~\cite{yu2019does} that train two networks simultaneously, where in each mini-batch, it selects small-loss samples to be used in the training of the other model.
JoCoR~\cite{wei2020combating} improves upon Co-teaching+ by using a contrastive loss to jointly train both models. 
DivideMix~\cite{li2020dividemix} has advanced the area with a similar combination of sample selection and prediction disagreement using semi-supervised learning, co-teaching and small-loss detection with a Gaussian Mixture Model (GMM). InstanceGM~\cite{garg2022instance} combines graphical model with DivideMix to achieve promising results.
These methods show that sample selection based on the small-loss assumption is one of the core components for achieving SOTA  performance. 
However, the small loss signal used to select samples is poorly integrated with prediction disagreement since both models will quickly converge to produce similar loss values for all training samples, resulting in little disagreement between models, which increases the risk of confirmation bias.

\textbf{Transition matrix} methods aim to estimate a noise transition matrix to guarantee that the classifier learned from the noisy data is consistent with the optimal classifier~\cite{xia2019anchor,patrini2017making,cheng2022instance} 
F-correction~\cite{patrini2017making} uses a two-step solution to heuristically estimate the noise transition matrix. T-revision~\cite{xia2019anchor} argues that anchor points are not necessary for estimating the transition matrix and proposes a solution for selecting reliable samples to replace anchor points. kMEIDTM~\cite{cheng2022instance} proposes an anchor-free method for estimating instance-dependent transition matrix by applying manifold regularization during the training. 
The main issue with the methods above is that it is challenging to estimate the transition matrix accurately, particularly an instance-dependent transition matrix that contains little support from the training set.  Furthermore, real-world scenarios often contain  out-of-distribution samples that are hard to represent in the transition matrix. 

\textbf{Multi-view learning} (MVL) studies the integration of knowledge from different views of the data to capture consensus and complementary information across different views. Traditional MVL methods~\cite{blum1998combining,sindhwani2005co} aimed to encourage the convergence of patterns from different views. 
For example, Co-training~\cite{blum1998combining} uses two views of web-pages (i.e., text and hyperlinks on web-pages) to allow the use of inexpensive unlabelled data to augment a small labelled data. 
Considering that the quality and importance of different views could vary for real-world applications, recent methods~\cite{han2021trusted} weight the contribution of each view based on the estimated uncertainty.
In our paper, we explore this multi-view learning strategy to select clean and noisy samples and to iteratively re-label training samples, where the views are represented by the training labels, and the predictions by the two models that are trained using different learning strategies.

\vspace{-0.2cm}
\section{Method}
\label{sec:method}

\subsection{Problem Definition}

We denote the noisy training set as $\mathcal{D} = \{ (\mathbf{x}_i,\tilde{\mathbf{y}}_i)\}_{i=1}^{|\mathcal{D}|}$, where $\mathbf{x}_i \in \mathcal{X} \subset \mathbb{R}^{H \times W \times C}$ is the input image of size $H \times W$ with $C$ colour channels, and $\tilde{\mathbf{y}}_i  \in \mathcal{Y} \subset \{0,1\}^{|\mathcal{Y}|}$ is the one-hot (or multi-class) label representation. The goal of is to learn the \textbf{classification net} $n_{\theta}:\mathcal{X} \to \mathcal{L}$, parameterised by $\theta \in \Theta$, that outputs the logits $\mathbf{l} \in \mathcal{L} \subset \mathbb{R}^{|\mathcal{Y}|}$ for an image $\mathbf{x} \in \mathcal{X}$. 
Following the prediction-disagreement strategy, we also define the \textbf{reference net} denoted by $r_{\phi}:\mathcal{X} \to \mathcal{L}$, parameterised by $\phi  \in \Phi$, to be jointly trained with $n_{\theta}(.)$. 

\begin{figure}
    \centering
    \includegraphics[width=0.8\linewidth]{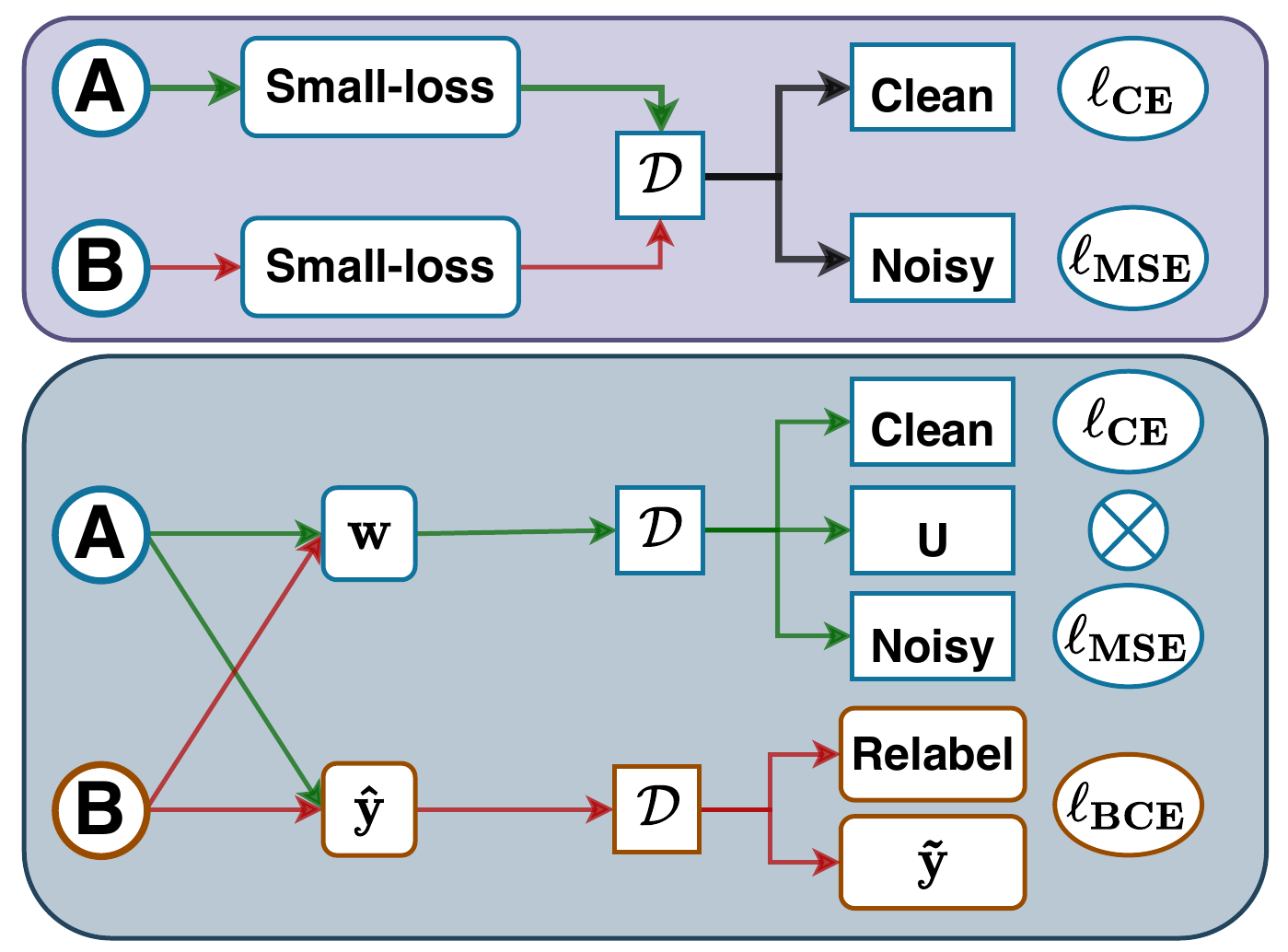}
    \caption{Comparison between traditional small-loss sample selection (top) and our AsyCo, consisting of prediction disagreement between the multi-class model A and multi-label model B (bottom). Traditional methods utilises the small-loss assumption for classifying samples as clean or noisy, 
    while our multi-view sample selection uses prediction disagreements to update the sample-selection variable $\mathbf{w}$ for classifying samples as clean, noisy or unmatched (U) to train the classification net A. Our multi-view re-labelling  selects ambiguous samples and maximise disagreement by updating the re-labelling variable $\mathbf{\hat{y}}$ for training the reference net B.}
    \label{fig:framework2}
\end{figure}

AsyCo\footnote{Algorithm in supplementary material.%
} is based on alternating the training of the multi-class model $n_{\theta}(.)$ and the multi-label model $r_{\phi}(.)$, which allows the formation of three label views for the training samples $\{\mathbf{x}_i\}_{i=1}^{|\mathcal{D}|}$: 1) the original training label $\tilde{\mathbf{y}}_i$, 2) the classification net multi-class prediction $\tilde{\mathbf{y}}^{(n)}_i$,  and 3) the reference net multi-label prediction $\tilde{\mathbf{y}}^{(r)}_i$.
Using these views, we introduce new methods to estimate the sample-selection variable $\mathbf{w}$ that classifies training samples into clean or noisy, and the re-labelling  variable $\hat{\mathbf{y}}$ that holds multiple top-ranked labels for training samples, where $\mathbf{w}$ is used for training the multi-class model $n_{\theta}(.)$, and $\hat{\mathbf{y}}$ for training the multi-label model $r_{\phi}(.)$. 
\cref{fig:framework2} depicts AsyCo, in comparison with prediction disagreement methods based on co-teaching and small-loss sample selection.

\subsection{Asymmetric Co-teaching Optimisation}
\label{sec:optimiz}

Our Asymmetric co-teaching optimisation trains a multi-class model with the usual cross-entropy (CE), but the other model is trained with multi-label learning ~\cite{ridnik2021asymmetric} that associates samples with multiple labels and utilises binary cross-entropy (BCE) to train for each label independently.
We have two goals with the multi-label model: 1) maximise the disagreement with the multi-class model, and 2) formulate a mechanism to find the most likely clean labels by selecting multiple top-ranked labels of training samples.
While the first goal is motivated by the training strategy differences, the second goal is motivated by the hypothesis that a possible cause of the overfitting of noisy labels is the single-class constraint that forces multi-class models to fit only one class. By removing this constraint, the true clean label is likely to be within the top-ranked candidate labels\footnote{Training strategy visualization in supplementary material.}.
Our AsyCo optimisation starts with a warmup stage of supervised learning to train both networks with:
\begin{equation}
    \begin{split}
        \theta^{\dagger} &= \arg\min_{\theta}
        \sum_{(\mathbf{x}_i,\tilde{\mathbf{y}}_i)  \in \mathcal{D}}\ell_{\mathrm{CE}}(\tilde{\mathbf{y}}_i,\sigma_{sm}(n_{\theta}(\mathbf{x}_i))), \\
        \phi^{\dagger} &= \arg\min_{\phi}
        \sum_{(\mathbf{x}_i,\tilde{\mathbf{y}}_i)  \in \mathcal{D}} \ell_{\mathrm{BCE}}(\tilde{\mathbf{y}}_i,\sigma_{sg}(r_{\phi}(\mathbf{x}_i))),
    \end{split}
    \label{eq:warmup_loss}
\end{equation}
where  $\sigma_{sm}(.)$ and $\sigma_{sg}(.)$ are the softmax and sigmoid activation functions, respectively, $\ell_{\mathrm{CE}}(.)$ represents the CE loss for multi-class learning, and $\ell_\mathrm{BCE}$ denotes the BCE loss for multi-label learning. The two models from~\eqref{eq:warmup_loss} will provide predictions as follows:
\begin{equation}
    \begin{split}
        \tilde{\mathbf{y}}_i^{(n)} &= \mathrm{OneHot}( n_{\theta^{\dagger}}(\mathbf{x}_i)),\\
        \tilde{\mathbf{y}}_i^{(r)} &= \mathrm{TopK}( r_{\phi^{\dagger}}(\mathbf{x}_i)),
    \end{split}
    \label{eq:prediction}
\end{equation}
where $\tilde{\mathbf{y}}_i^{(n)} \in \mathcal{Y}$ is the one-hot single-label prediction by $n_{\theta^{\dagger}}(\mathbf{x}_i)$, and $\tilde{\mathbf{y}}_i^{(r)} \in \{0,1\}^{|\mathcal{Y}|}$ is the top-$K$ multi-label prediction of $r_{\phi^{\dagger}}(\mathbf{x}_i)$ (i.e., the largest $K$ values from $r_{\phi^{\dagger}}(.)$ will set $\tilde{\mathbf{y}}_i^{(r)}$ to $1$ and the rest are set to $0$). However, removing the single-class constraint from multi-class classification inevitably weakens the model performance. Thus, we aim to extract useful information from top-ranked candidate labels to help training $n_\theta$ with multi-view consensus, explained below, which uses the label views produced by the predictions from $n_\theta$ and $r_\phi$ and the training labels, to select samples for training $n_\theta$ and re-label samples for training $r_\phi$.

\subsection{Multi-view Consensus}

One of the objectives of 
maximising prediction disagreement between models is to improve  sample selection accuracy for co-teaching.
We propose a new sample selection based on multi-view consensus, where each sample $\mathbf{x}_i$ has three label views: the single-label training label $\tilde{\mathbf{y}}_i$, the single-label one-hot prediction $\tilde{\mathbf{y}}_i^{(n)}$, and the multi-label top-$K$ prediction $\tilde{\mathbf{y}}_i^{(r)}$.
These multiple views allow us to build  training subsets given prediction disagreements, as shown in Tab.~\ref{tab:agreement_labels}, where the  Agreement Degree (AG) score is defined as:
\begin{equation}
    \text{AG}(\tilde{\mathbf{y}},\tilde{\mathbf{y}}^{(n)},\tilde{\mathbf{y}}^{(r)}) = \tilde{\mathbf{y}}^{\top}\tilde{\mathbf{y}}^{(n)} +
          {\tilde{\mathbf{y}}^{(n)}}^{\top} \tilde{\mathbf{y}}^{(r)}
          + \tilde{\mathbf{y}}^{\top} \tilde{\mathbf{y}}^{(r)}
    \label{eq:agreement_degree}
\end{equation}

\begin{table}[t!]
\caption{Three possible label views: the training label $\tilde{\mathbf{y}}_i$, the single-label one-hot prediction $\tilde{\mathbf{y}}_i^{(n)}$, and the multi-label top-$K$ prediction $\tilde{\mathbf{y}}_i^{(r)}$. The combination of these multiple views form the subsets, defined in the first column, with agreement scores $\text{AG}(.)$, from~\eqref{eq:agreement_degree}, in the last column. %
}
\centering
\scalebox{0.9}{
\begin{tabular}{@{}c|c|c|c||c}
\toprule \hline
           Subsets & $\tilde{\mathbf{y}}^{\top}\tilde{\mathbf{y}}^{(n)}$ & 
          ${\tilde{\mathbf{y}}^{(n)}}^{\top} \tilde{\mathbf{y}}^{(r)}$ 
          & $\tilde{\mathbf{y}}^{\top} \tilde{\mathbf{y}}^{(r)}$ &
          $\text{AG}(.)$ \\ \hline
Core (C)      & 1                                                    & 1                                                                & 1         & 3                                                           \\ 
Side-Core (SC) & 0                                                   & 1                                                                & 1                 & 2                                                   \\ \hline
NY        & 1                                                    & 0                                                               & 0              & 1                                                     \\
NR        & 0                                                   & 1                                                                & 0                  & 1                                                 \\
RY        & 0                                                   & 0                                                               & 1               & 1                                                     \\ \hline
Unmatched (U) & 0                                                   & 0                                                               & 0                  & 0                                         \\ \hline \bottomrule        
\end{tabular}
}
\label{tab:agreement_labels}
\end{table}

The \textbf{training of the classification net $n_{\theta}(.)$}
has the goals of producing the testing model and of maximising the disagreement with $r_{\phi}(.)$.
This training employs a semi-supervised learning strategy~\cite{berthelot2019mixmatch}, which requires the division of the training set into clean and noisy sets.
Unlike previous methods that rely on the small-loss assumption to classify training samples into clean or noisy~\cite{li2020dividemix,han2018co,arazo2019unsupervised}, we utilize the subsets created by prediction disagreements from the multiple label views shown in Tab.~\ref{tab:agreement_labels}. 
For training $n_{\theta}(.)$, we first discard all samples in the subset $\mathrm{Unmatched}$ given their high level of uncertainty because both models disagree with each other and with the training label. 
For the remaining samples, we seek label agreements between pair of views beyond its own prediction.
More specifically, training samples are classified as clean when $\tilde{\mathbf{y}}^{\top}\tilde{\mathbf{y}}^{(r)} = 1$, which indicates that the training label matches one of the top ranked predictions by $r_{\phi}(.)$. 
Such agreement from label views $\tilde{\mathbf{y}}$ and $\tilde{\mathbf{y}}^{(r)}$ indicates that the training label $\tilde{\mathbf{y}}$ is within the top-ranked predictions by $r_{\phi}(.)$, but may not match the prediction by $n_{\theta}(.)$. Therefore, classifying such samples as clean can help maximise the disagreement with $r_\phi$ and alleviate confirmation bias.
The remaining samples with $\tilde{\mathbf{y}}^{\top}\tilde{\mathbf{y}}^{(r)} = 0$ are classified as noisy because of the insufficient support by $r_{\phi}(.)$ for the training label $\tilde{\mathbf{y}}$.
Therefore, based on the criterion described above, the classification net $n_\theta$ is trained with $\{\mathrm{C}, \mathrm{SC}, \mathrm{RY}\}$ as clean and $\{\mathrm{NY}, \mathrm{NR}\}$ as noisy, defined by the following sample-selection variable:
\begin{equation}
    \mathbf{w}_i =
\left\{\begin{array}{lll} 
+1, & \text{ if }
\text{AG}(\tilde{\mathbf{y}}_i,\tilde{\mathbf{y}}^{(n)}_i,\tilde{\mathbf{y}}^{(r)}_i) > 0\text{ and }\tilde{\mathbf{y}}_i^{\top}\tilde{\mathbf{y}}_i^{(r)} = 1,\\
0, & \text{ if }
\text{AG}(\tilde{\mathbf{y}}_i,\tilde{\mathbf{y}}^{(n)}_i,\tilde{\mathbf{y}}^{(r)}_i) > 0\text{ and }\tilde{\mathbf{y}}_i^{\top}\tilde{\mathbf{y}}_i^{(r)} = 0,\\
-1, & \text{ if }
\text{AG}(\tilde{\mathbf{y}}_i,\tilde{\mathbf{y}}^{(n)}_i,\tilde{\mathbf{y}}^{(r)}_i) = 0 ,
\end{array}
\right.	
    \label{eq:w}
\end{equation}
where $\mathbf{w}_i\in\{+1,0,-1\}$ denotes a clean, noisy, and unmatched training sample, respectively.

The training of $n_{\theta}(.)$ is performed by %
\begin{equation}
\begin{split}
    \theta^{*} & = \arg\min_{\theta}
    \sum_{\substack{(\mathbf{x}_i,\tilde{\mathbf{y}}_i) \in \mathcal{D}\\\mathbf{w}_i=+1}} \ell_{CE}(\tilde{\mathbf{y}}_i, \sigma_{sm} (n_{\theta}(\mathbf{x}_i))) \\
    &+ \lambda \sum_{\substack{(\mathbf{x}_i,\tilde{\mathbf{y}}_i) \in \mathcal{D}\\ \mathbf{w}_i=0}} \ell_{MSE}( \upsilon(\sigma_{sm}(n_{\theta}(\mathbf{x}_i)),T), \sigma_{sm}(n_{\theta}(\mathbf{x}_i))),
\end{split}
\label{eq:loss_training_net}
\end{equation}
where $\upsilon(.,T)$ is a sharpening function~\cite{li2020dividemix} parameterised by the temperature $T$, and $\lambda$ is the weight to control the strength of the unsupervised learning with the noisy labels, and $\ell_{MSE}(.)$ denotes the mean square error loss function.

The \textbf{training of the reference net $r_\phi(.)$} has the goals of maximising the disagreement with $n_\theta(.)$ using the multi-view consensus from~\cref{tab:agreement_labels}, and maintaining the top-ranked labels of training samples as clean label candidates.
To achieve that, we focus on designing a new supervisory training signal by re-labelling the samples where predictions by $n_{\theta}(.)$ and $r_{\phi}(.)$ match (i.e., ${\tilde{\mathbf{y}}^{(n)}}^{\top}\tilde{\mathbf{y}}^{(r)}=1$) and the prediction by $n_{\theta}(.)$  does not match the training label $\tilde{\mathbf{y}}$ (i.e.,  ${\tilde{\mathbf{y}}}^{\top}\tilde{\mathbf{y}}^{(n)}=0$). 
The training samples that meet this  condition can be regarded as hard to fit by $n_{\theta}(.)$,
with the top-ranked predictions by $\tilde{\mathbf{y}}^{(r)}$ being likely to contain the hidden clean label.
The conditions above indicates that we select samples from $\mathrm{SC} \bigcup \mathrm{NR}$ from~\cref{tab:agreement_labels} for re-labelling.
For samples in $\mathrm{SC}$, since $n_{\theta}(.)$ is trained with supervised learning in~\eqref{eq:loss_training_net}, the maximisation of prediction disagreement is achieved by re-labelling the sample to $\tilde{\mathbf{y}}^{(n)}$. 
For samples in $\mathrm{NR}$, $n_{\theta}(.)$ is trained with unsupervised learning in~\eqref{eq:loss_training_net}, so the prediction disagreement is maximised by re-labelling the sample to $\tilde{\mathbf{y}} + \tilde{\mathbf{y}}^{(n)}$, forming a multi-label target.
We define the re-labelling variable $\hat{\mathbf{y}}$ to represent the new supervisory training signal, as follows:
\begin{equation}
    \hat{\mathbf{y}}_i =
\left\{\begin{array}{lll} 
\tilde{\mathbf{y}}_i^{(n)}, & \text{ if } (\mathbf{x}_i,\tilde{\mathbf{y}}_i) \in  \mathrm{SideCore},\\
\tilde{\mathbf{y}}_i + \tilde{\mathbf{y}}_i^{(n)}, & \text{ if } (\mathbf{x}_i,\tilde{\mathbf{y}}_i) \in \mathrm{NR},\\
\tilde{\mathbf{y}}_i, & \text{otherwise},
\end{array}
\right.	
    \label{eq:haty}
\end{equation}
with training of $r_{\phi}(.)$  achieved with:
\begin{equation}
    \phi^{*} = \arg\min_{\phi} \sum_{i=1}^{|\mathcal{D}|}  \ell_{BCE}(\hat{\mathbf{y}}_i,\sigma_{sg}(r_{\phi}(\mathbf{x}_i))).
    \label{eq:training_reference_net}
\end{equation}

Note that this re-labelling is iteratively done at every epoch. The testing procedure depends exclusively on the classification net $n_{\theta}(.)$.

\section{Experiments}

We show the results of extensive experiments on instance-dependent synthetic noise benchmarks with datasets CIFAR10 and CIFAR100~\cite{krizhevsky2009learning} with various noise rates and on three real-world datasets, namely: Animal-10N~\cite{song2019selfie}, Red Mini-ImageNet~\cite{jiang2020beyond} and Clothing1M~\cite{xiao2015learning}.

\subsection{Datasets}

\textbf{CIFAR10/100}. For CIFAR10 and CIFAR100~\cite{krizhevsky2009learning}, the training set contains 50K images and testing set contains 10K images of size 32 $\times$ 32 $\times$ 3. CIFAR10 has 10 classes and CIFAR100 has 100 classes. We follow previous work~\cite{xia2020part} for generating instance-dependent noise with rates in \{0.2, 0.3, 0.4, 0.5\}.
\textbf{Red Mini-ImageNet} is proposed by ~\cite{jiang2020beyond} based on Mini-ImageNet~\cite{deng2009imagenet}. The images and their corresponding labels are annotated by Google Cloud Data Labelling Service. This dataset is proposed to study real-world web-based noisy label. 
Red Mini-ImageNet has 100 classes with each class containing 600 images from ImageNet. The images are resized to 32 $\times$ 32 from the original 84 $\times$ 84 pixels to allow a fair comparison with other baselines~\cite{xu2021faster, jiang2020beyond}. We test our method on noise rates in \{20\%, 40\%, 60\%, 80\%\}.
\textbf{Animal 10N} is a real-world dataset proposed in~\cite{song2019selfie}, which contains 10 animal species with similar appearances (wolf and coyote, hamster and guinea pig, etc.). The training set size is 50K and testing size is 10K, where we follow the same setup as~\cite{song2019selfie}.
\textbf{Clothing 1M} is a real-world dataset with 100K images and 14 classes. The labels are generated from surrounding text with an estimated noise ratio of 38.5\%. We follow a common setup using a training image size of 224 $\times$ 224 pixels. The dataset also contains clean training, clean validation and clean test sets with 50K, 14K and 10K images. We do not use clean training and clean validation, only the clean testing is  used for measuring model performance.

\begin{table*}[t!]
\centering
\scalebox{0.95}{
\begin{tabular}{l|cccc|cccc}
\toprule \hline
\multirow{2}{*}{Methods} & \multicolumn{4}{c|}{CIFAR10}              & \multicolumn{4}{c}{CIFAR100}  \\ \cline{2-9} 
                         & 0.2   & 0.3        & 0.4     & 0.5        & 0.2   & 0.3   & 0.4   & 0.5   \\ \hline
CE                       & 75.81 & 69.15      & 62.45   & 39.42      & 30.42 & 24.15 & 21.34 & 14.42 \\
Mixup~\cite{zhang2017mixup}                    & 73.17 & 70.02      & 61.56   & 48.95      & 32.92 & 29.76 & 25.92 & 21.31 \\
Forward~\cite{patrini2017making}                  & 74.64 & 69.75      & 60.21   & 46.27      & 36.38 & 33.17 & 26.75 & 19.27 \\
T-Revision~\cite{xia2019anchor}              & 76.15 & 70.36      & 64.09   & 49.02      & 37.24 & 36.54 & 27.23 & 22.54 \\
Reweight~\cite{liu2015classification}                & 76.23 & 70.12      & 62.58   & 45.46      & 36.73 & 31.91 & 28.39 & 20.23 \\
PTD-R-V~\cite{xia2020part}                  & 76.58 & 72.77      & 59.50   & 56.32      & 65.33 & 64.56 & 59.73 & 56.80 \\
Decoupling~\cite{malach2017decoupling}               & 78.71 & 75.17      & 61.73   & 50.43      & 36.53 & 30.93 & 27.85 & 19.59 \\
Co-teaching~\cite{han2018co}              & 80.96 & 78.56      & 73.41   & 45.92      & 37.96 & 33.43 & 28.04 & 23.97 \\
MentorNet~\cite{jiang2018mentornet}                & 81.03 & 77.22      & 71.83   & 47.89      & 38.91 & 34.23 & 31.89 & 24.15 \\
CausalNL~\cite{yao2021instance}                 & 81.79 & 80.75      & 77.98   & 78.63      & 41.47 & 40.98 & 34.02 & 32.13 \\
CAL~\cite{zhu2021second}                      & 92.01 & -          & 84.96   & -          & 69.11 & -     & 63.17 & -     \\
kMEIDTM~\cite{cheng2022instance}                  & 92.26 & 90.73      & 85.94   & 73.77      & 69.16 & 66.76 & 63.46 & 59.18 \\ \hline
DivideMix~\cite{li2020dividemix} $\theta^{(1)} $  test  $^\dagger$      & 94.62 & 94.49      & 93.50    & 89.07      & 74.43 & 73.53 & \color{purple}{69.18} & 57.52 \\
Ours              & \color{purple}{96.00} & \color{purple}{95.82}      & \color{purple}{95.01}   & \color{purple}{94.13}      & \color{purple}{76.02} & \color{purple}{74.02} & 68.96 & \color{purple}{60.35} \\ \hline
DivideMix ~\cite{li2020dividemix} $^\dagger$              & 94.80 & 94.60      & 94.53   & 93.04      & 77.07 & 76.33 & 70.80 & 58.61 \\
Ours 2$\times n_\theta$ test              & \color{teal}{96.56} & \color{teal}{96.11}      & \color{teal}{95.53}   & \color{teal}{94.86}      & \color{teal}{78.50} & \color{teal}{77.32} & \color{teal}{73.32} & \color{teal}{65.96} \\ \hline \bottomrule
\end{tabular}
}

\caption{Test accuracy (\%) of different methods on CIFAR10/100 with instance-dependent noise~\cite{xia2020part}. Results reproduced from publicly available code are presented with $\dagger$. Best single/ensemble inference results are labelled with \color{purple}{red}/\color{teal}{green}.}
\label{tab:cifar}
\end{table*}
\subsection{Implementation}
For CIFAR10/10 and Red Mini-ImageNet we use Preact-ResNet18~\cite{he2015deep} and train it for 200 epochs with SGD with momentum=0.9, weight decay=5e-4 and batch size=128. The initial learning rate is 0.02 and reduced by a factor of 10 after 150 epochs. The warmup period for all three datasets is 10 epochs. We set $\lambda=25$ in~\eqref{eq:loss_training_net} for CIFAR10 and Red Mini-ImageNet, and $\lambda=100$ for CIFAR100. 
In~\eqref{eq:prediction}, we set $K=1$  for CIFAR10 and $K=3$ for CIFAR100 and Red Mini-ImageNet. These values are fixed for all noise rates. For data augmentations, we use random cropping and random horizontal flipping for all three datasets.

For Animal 10N, we follow a common setup used by previous methods with a VGG-19BN~\cite{simonyan2014very} architecture, trained for 100 epochs with SGD with momentum=0.9, weight decay=5e-4 and batch size=128. The initial learning rate is 0.02, and reduced by a factor of 10 after 50 epochs. The warmup period is 10 epochs.
We set $\lambda=25$ and $K=2$. For data augmentations, we use random cropping and random horizontal flipping.

For Clothing1M, we use ImageNet~\cite{deng2009imagenet} pre-trained ResNet50~\cite{he2015deep} and train it for 80 epochs with SGD with momentum=0.9, weight decay=1e-3 and batch size=32. The warmup period is 1 epoch. The initial learning rate is set to 0.002 and reduced by a factor of 10 after 40 epochs. Following DivideMix~\cite{li2020dividemix}, we also sample 1000 mini-batches from the training set to ensure the training set is pseudo balanced. We set $K=4$. For data augmentation, we first resize the image to 256 $\times$ 256 pixels, then random crop to 224 $\times$ 224 and random horizontal flipping.

For the semi-supervised training of $n_\theta(.)$, we use  MixMatch~\cite{berthelot2019mixmatch}  from DivideMix~\cite{li2020dividemix}. 
We also extend our method to train two $n_\theta(.)$ models and use ensemble prediction at inference time, similarly to DivideMix~\cite{li2020dividemix}. 
We denoted this variant as $2\times n_\theta$. Our code is implemented in Pytorch~\cite{paszke2019pytorch} and all experiments are performed on an RTX 3090\footnote{Time of Different sample selection comparison in supplementary.}

\subsection{Comparison with SOTA Methods}
We compare our AsyCo with the following methods: 
1) CE, which trains the classification network with standard CE loss on the noisy dataset; 
2) Mixup~\cite{zhang2017mixup}, which employs mixup on the noisy dataset; 
3) Forward~\cite{patrini2017making}, which estimates the noise transition matrix in a two-stage training pattern; 
4) T-Revision~\cite{xia2019anchor}, which finds reliable samples to replace anchor points for estimating transition matrix; 
5) Reweight~\cite{liu2015classification}, which utilizes a class-dependent transition matrix to correct the loss function; 
6) PTD-R-V~\cite{xia2020part}, which proposes a part-dependent transition matrix for accurate estimation; 
7) Decoupling~\cite{malach2017decoupling}, which trains two networks on samples whose predictions from the network are different; 
8) Co-teaching~\cite{han2018co}, which trains two networks and select small-loss samples as clean samples; 
9) MentorNet~\cite{jiang2018mentornet}, which utilizes a teacher network for selecting noisy samples; 
10) CausalNL~\cite{yao2021instance}, which discovers a causal relationship in noisy dataset and combines it with Co-Teaching; 
11) CAL~\cite{zhu2021second}, which uses second-order statistics with a new loss function; 12) kMEIDTM~\cite{cheng2022instance}, which learns instance-dependent transition matrix by applying manifold regularization during the training; 
13) DivideMix~\cite{li2020dividemix}, which combines semi-supervised learning, sample selection and Co-Teaching to achieve SOTA results; 
14) FaMUS~\cite{xu2021faster}, which is a meta-learning method that learns the weight of training samples to improve the meta-learning update process;
15) Nested~\cite{chen2021boosting}, which is a novel feature compression method that uses nested dropout to regularize features when training with noisy label--this approach can be combined with existing techniques such as Co-Teaching~\cite{han2018co}; and 
16) PLC~\cite{zhang2021learning}, which is a method that produces soft pseudo label when learning with label noise.

\subsection{Experiment Results}
\label{sec:CIFAR_results}

\textbf{Synthetic Noise Benchmarks}. The experimental results of our proposed AsyCo with instance-dependent noise on CIFAR10/100 are shown in Tab.~\ref{tab:cifar}. 
We reproduce DivideMix~\cite{li2020dividemix} in this setup with single model at inference time denoted by $\theta^{(1)}$ and also the original ensemble inference. 
Compared with the best baselines, our method achieves large improvements for all noise rates. 
On CIFAR10, we achieve $\approx1.5\%$ improvements for low noise rates and $\approx1\%$ to $5\%$ improvements for high noise rates. 
For CIFAR100, we improve between $\approx1.5\%$ and $\approx7\%$ for many noise rates. 
Note that our result is achieved without using small-loss sample selection, which is a fundamental technique for most noisy label learning methods~\cite{li2020dividemix,han2018co,jiang2018mentornet}. 
The superior performance of AsyCo indicates that our multi-view consensus for sample selection and top-rank re-labelling are effective when learning with label noise.

\begin{table}[t!]
\centering
\scalebox{0.9}{
\begin{tabular}{l|cccc}
\toprule \hline
\multirow{2}{*}{Method} & \multicolumn{4}{c}{Noise rate}                                                                                \\ \cline{2-5} 
                        & 0.2                       & 0.4                       & 0.6                       & 0.8                       \\ \hline
CE                      & 47.36                     & 42.70                     & 37.30                     & 29.76                     \\
Mixup~\cite{zhang2017mixup}                   & 49.10                     & 46.40                     & 40.58                     & 33.58                     \\
DivideMix~\cite{li2020dividemix}               & 50.96                     & 46.72                     & 43.14                     & 34.50                     \\
MentorMix~\cite{jiang2020beyond}               & 51.02                     & 47.14                     & 43.80                     & 33.46                     \\
FaMUS~\cite{xu2021faster}                   & 51.42                     & 48.06                     & 45.10                     & 35.50                     \\ \hline
Ours                    & \color{purple}{59.40}                     & \color{purple}{55.08}                     & \color{purple}{49.78}                     & \color{purple}{41.02}                    \\
Ours 2$\times n_\theta$ test                    & \color{teal}{61.98} & \color{teal}{57.46} & \color{teal}{51.86} & \color{teal}{42.58} \\ \hline \bottomrule
\end{tabular}}
\caption{Test accuracy (\%) of different methods on Red Mini-ImageNet with different noise rates. Baselines results are from FaMUS~\cite{xu2021faster}. Best results with single/ensemble inferences are labelled with \color{purple}{red}/\color{teal}{green}. }
\label{tab:red}
\end{table}

\begin{table}[t!]
\centering 
\scalebox{0.9}{
\begin{tabular}{c|c}
\toprule \hline
Method                  & Accuracy \\ \hline
CE                      & 79.4     \\
Nested~\cite{chen2021boosting}          & 81.3     \\
Dropout + CE  ~\cite{chen2021boosting}           & 81.1     \\
SELFIE  ~\cite{song2019selfie}                & 81.8     \\
PLC ~\cite{zhang2021learning}                   & 83.4     \\
Nested + Co-Teaching  ~\cite{chen2021boosting}    & 84.1     \\ \hline
Ours                    & \color{purple}{85.6}     \\ 
Ours 2$\times n_\theta$ & \color{teal}{86.3%
}        \\ \hline \bottomrule
\end{tabular}}
\caption{Test accuracy (\%) of different methods on Animal-10N. Baselines results are presented with Nested Dropout~\cite{chen2021boosting}. Best  single/ensemble inference results are labelled with \color{purple}{red}/\color{teal}{green}. }
\label{tab:animal}
\end{table}

\begin{table*}[t!]
\centering
\scalebox{0.9}{
\begin{tabular}{c|l|ccccc|c}
\toprule \hline
\multirow{2}{*}{Single}   & Methods  & CE          & Forward  ~\cite{patrini2017making}    & PTD-R-V~\cite{xia2020part} & ELR ~\cite{liu2020early}     & kMEIDTM ~\cite{cheng2022instance}  & Ours                    \\ \cline{2-8} 
                          & Accuracy & 68.94       & 69.84        & 71.67   & 72.87    & 73.34     & \color{purple}{73.60}                    \\ \hline
\multirow{2}{*}{Ensemble} & Methods  & Co-Teaching~\cite{han2018co} & Co-Teaching+~\cite{yu2019does} & JoCoR~\cite{wei2020combating}   & CausalNL~\cite{yao2021instance} & DivideMix~\cite{li2020dividemix} & Ours 2$\times n_\theta$ \\ \cline{2-8} 
                          & Accuracy & 69.21       & 59.3         & 70.3    & 72.24    & \color{teal}{74.60}      & \color{teal}{74.43}       \\ \hline \bottomrule            
\end{tabular}}
\caption{Test accuracy (\%) of different methods on Clothing1M. Best single/ensemble inference results are labelled with \color{purple}{red}/\color{teal}{green}. }
\label{tab:clothing}
\end{table*}

\textbf{Real-world Noisy-label Datasets}. In~\cref{tab:red}, we present results on Red Mini-ImageNet~\cite{jiang2020beyond}.
Our method achieves SOTA results for all noise rates with 4\% to 8\% improvements in single model inference and 7\% to 10\% in ensemble inference. The improvement is significant compared with FaMUS~\cite{xu2021faster} with a gap of more than 6\%. Compared with DivideMix~\cite{li2020dividemix}, our method achieves between 6\% and 10\% improvements. In~\cref{tab:animal}, we present the results for Animal 10N~\cite{song2019selfie}, where the previous SOTA method was Nested Dropout + Co-Teaching~\cite{chen2021boosting}, which achieves 84.1\% accuracy. Our method achieves 85.6\% accuracy, which is 2.2\% higher than previous SOTA. Additionally, our ensemble version achieves 86.34\% accuracy, which improves 1\% more compared to our single inference model, yielding a new SOTA result.  
In~\cref{tab:clothing}, we show our result on Clothing1M~\cite{xiao2015learning}. In the single model setup, our model outperforms all previous SOTA methods. In the ensemble inference setup, our model shows comparable performance with the SOTA method DivideMix~\cite{li2020dividemix} and outperforms all other methods. Compared with other methods based on prediction disagreement~\cite{han2018co,yu2019does,wei2020combating}, our model improves by at least 3\%. The performance on these three real-world datasets indicates the superiority of our proposed AsyCo.

\section{Ablation Study}
\label{sec:ablation}

For the ablation study, we first visualise the training losses of subsets from~\cref{tab:agreement_labels} that are used by our multi-view consensus approach. We also  compare the accuracy of GMM selected clean samples and our multi-view selected samples. Then we test alternative approaches for multi-view sample selection and re-labelling. We perform all ablation experiments on the instance-dependent CIFAR10/100~\cite{xia2020part}.  

\begin{figure*}
        \centering
        \begin{subfigure}[b]{0.24\textwidth}
            \centering
            \includegraphics[width=\textwidth]{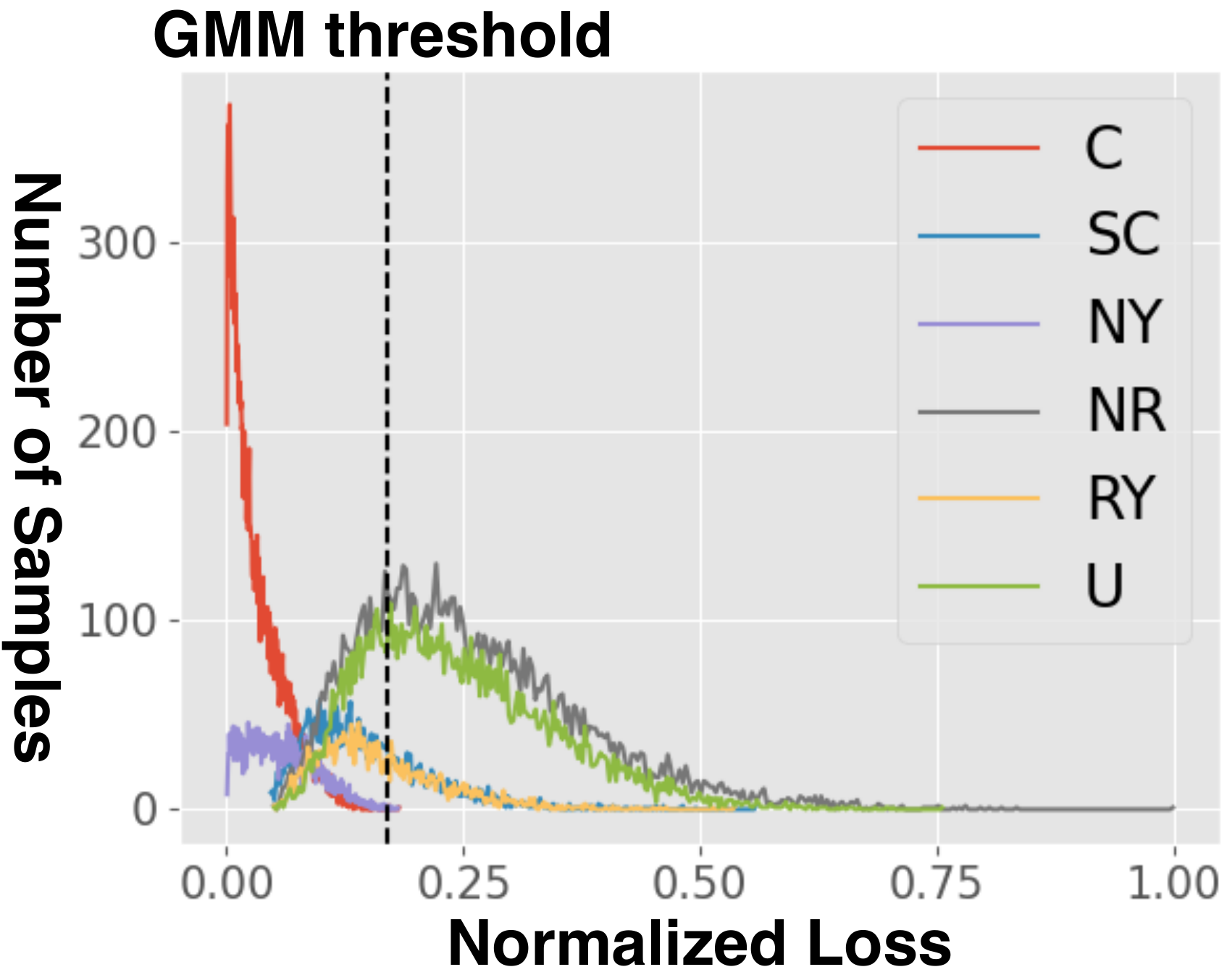}
            \caption[]%
            {CIFAR100 0.2 loss} 
            \label{fig:cifar100_0.2}
        \end{subfigure}
        \hfill
        \begin{subfigure}[b]{0.24\textwidth}  
            \centering 
            \includegraphics[width=\textwidth]{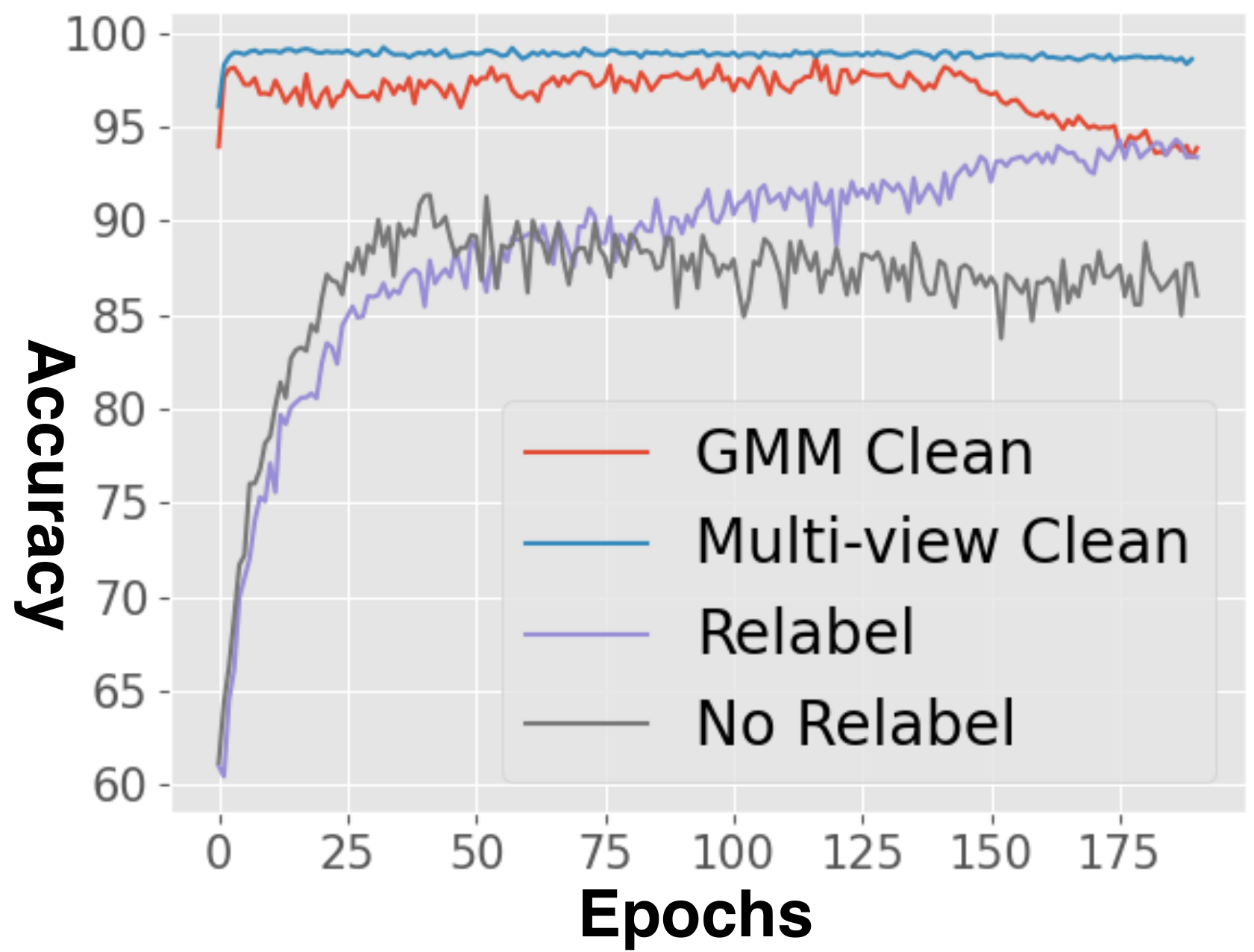}
            \caption[]%
            {CIFAR100 0.2 Accuracy}    
            \label{fig:acc_cifar100_0.2}
        \end{subfigure}
        \hfill
        \begin{subfigure}[b]{0.24\textwidth}   
            \centering 
            \includegraphics[width=\textwidth]{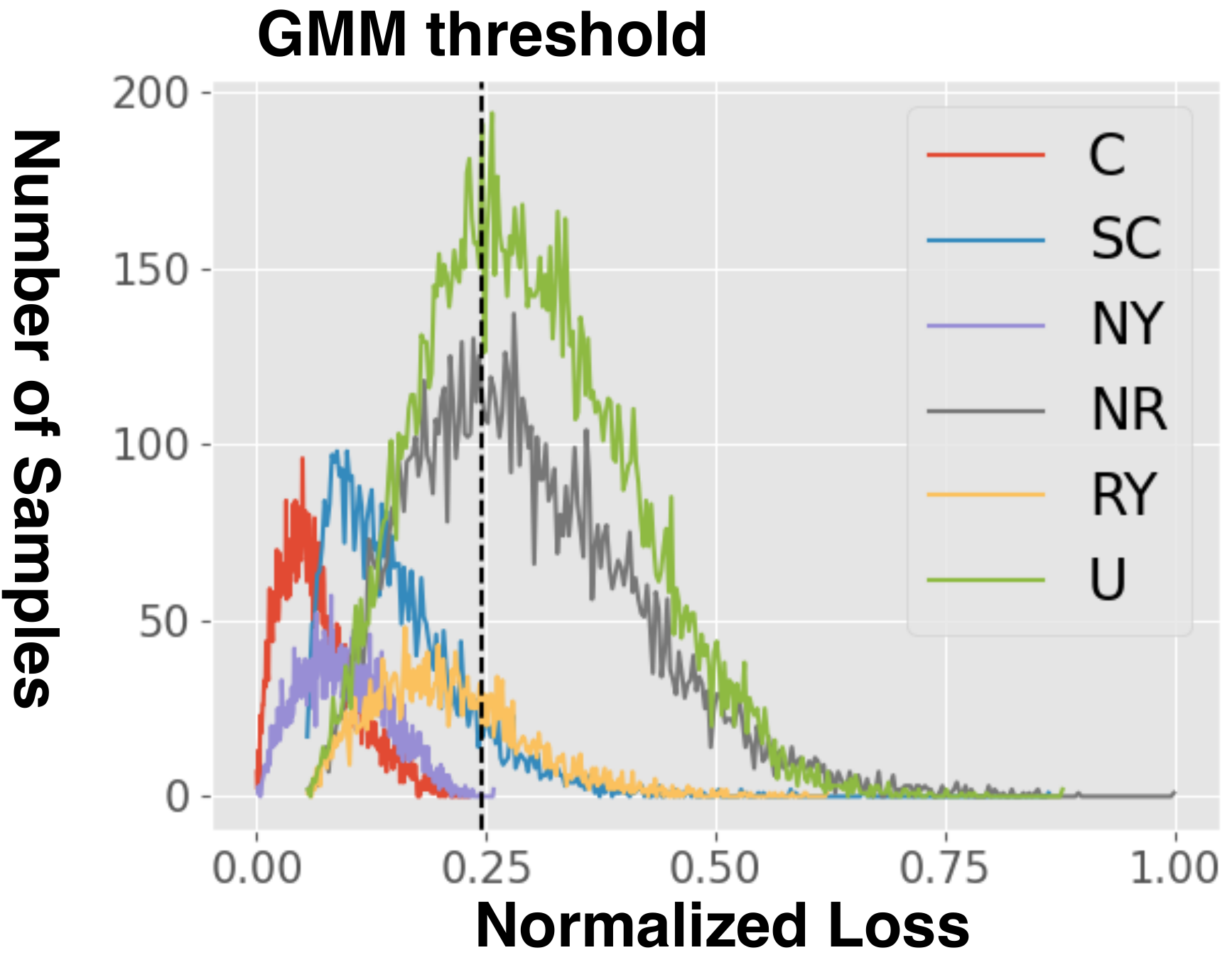}
            \caption[]%
             {CIFAR100 0.5 loss} 
             \label{fig:cifar100_0.5}
        \end{subfigure}
        \hfill
        \begin{subfigure}[b]{0.24\textwidth}
            \centering
            \includegraphics[width=\textwidth]{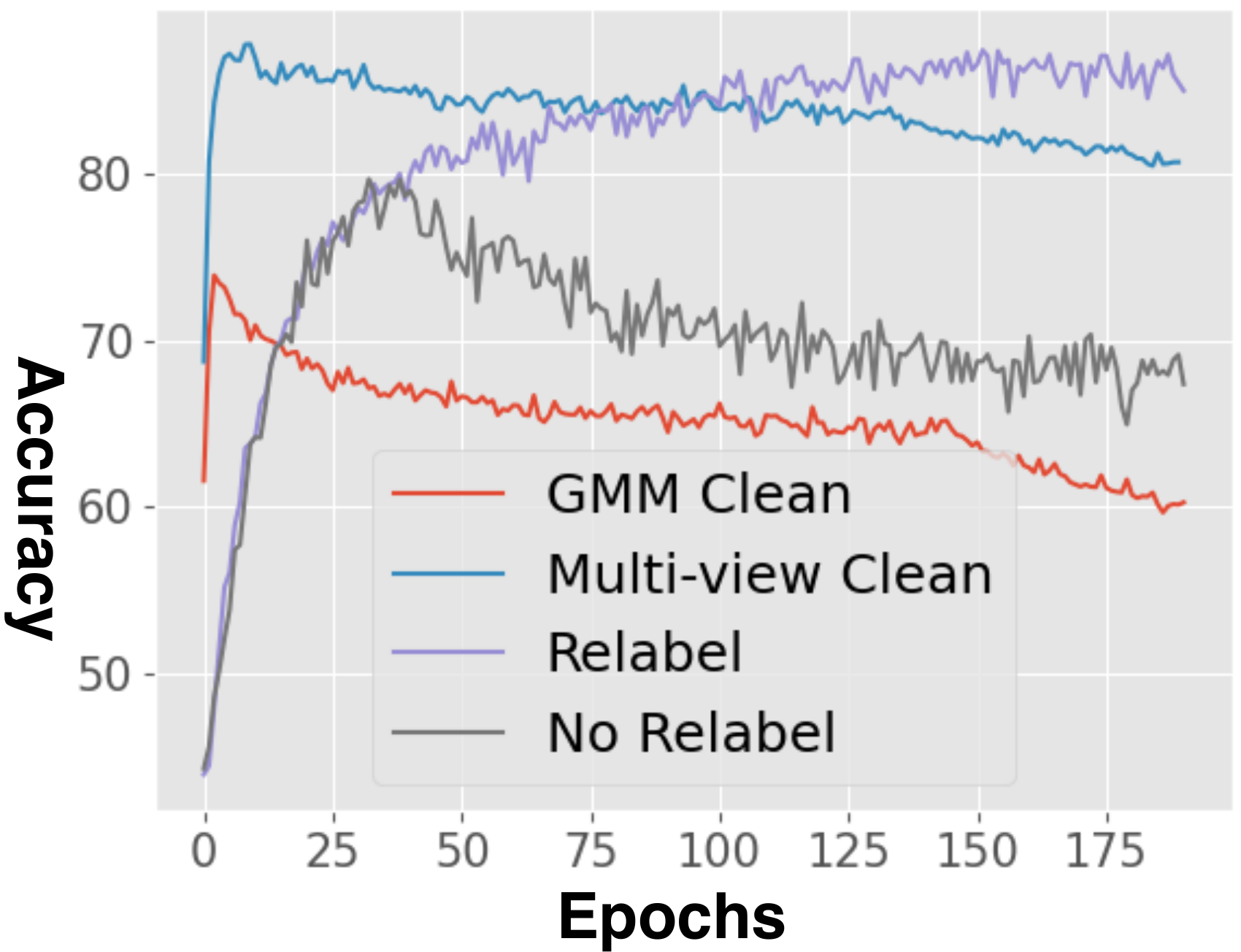}
            \caption[]%
             {CIFAR100 0.5 Accuracy} 
            \label{fig:acc_cifar100_0.5}
        \end{subfigure}
        \caption{ (a) and (c) are sample loss histograms for the subsets in Tab.~\ref{tab:agreement_labels} for CIFAR100 with 0.2 and 0.5 instance-dependent noise after warmup. Vertical dot line is GMM threshold. (b) and (d) are accuracy of clean set selected by GMM and our multi-view strategy. (b) and (d) also show accuracy of whether hidden clean labels within $r_\phi$ top-ranked prediction or not for multi-view re-labelling and not re-labelling.} 
        \label{fig:ablation_loss}
    \end{figure*}

\begin{table*}[t!]
\centering
\scalebox{0.9}{
\begin{tabular}{cl|cccc|cccc}
\toprule \hline
\multicolumn{1}{c|}{\multirow{2}{*}{Model}}      & \multirow{2}{*}{Ablation}                    & \multicolumn{4}{c|}{CIFAR10}                                     & \multicolumn{4}{c}{CIFAR100}                                      \\ \cline{3-10} 
\multicolumn{1}{c|}{}                            &                                              & 0.2           & 0.3            & 0.4            & 0.5            & 0.2            & 0.3            & 0.4            & 0.5            \\ \hline
\multicolumn{1}{c|}{\multirow{4}{*}{$n_\theta$}} & $\mathbf{w}_i = 0$ \text{ if } $(\mathbf{x}_i,\tilde{\mathbf{y}}_i) \in \mathrm{RY}$                                     & 93.28         & 93.85          & 92.54          & 82.60          & 73.58          & 71.51          & 65.51          & 56.65          \\
\multicolumn{1}{c|}{}                            & $\mathbf{w}_i = 0$ \text{ if } $(\mathbf{x}_i,\tilde{\mathbf{y}}_i) \in \mathrm{U}$                               & 95.71         & 94.88          & 94.34          & 91.60          &    75.10            &      72.64          &       67.42         &     57.55           \\
\multicolumn{1}{c|}{}                            & $\mathbf{w}_i = +1$ \text{ if } $(\mathbf{x}_i,\tilde{\mathbf{y}}_i) \in \mathrm{U}$ 
& 95.20         & 95.14          & 94.72          & 90.27          &   75.34             & 73.21               &      66.09          &     55.95           \\
\multicolumn{1}{c|}{}                            & Small-loss subsets                                & 92.37         & 91.80          & 90.93          & 78.53          & 70.10          & 69.52          & 64.69          & 56.35          \\ \hline

\multicolumn{1}{c|}{\multirow{4}{*}{$r_\phi$}}   & CE                                           & 95.22         & 94.83          & 83.48          & 64.96          & 73.33          & 69.29          & 63.82          & 54.83          \\

\multicolumn{1}{c|}{}                            & Frozen after warmup                                   & 91.19         & 88.97          & 84.72          & 67.57          & 68.73          & 65.36          & 58.88          & 48.13          \\
\multicolumn{1}{c|}{}                            & $\mathbf{\hat{y}_i} = \mathbf{\tilde{y}_i}$       & 95.42         & 94.69          & 90.53          & 84.95          & 74.43          & 71.75          & 62.25          & 53.69          \\
\multicolumn{1}{c|}{}                            & $\mathbf{\hat{y}_i} = \mathbf{\tilde{y}^{(n)}_i}$ & 94.29         & 94.23          & 94.13          & 93.67          & 74.55          & 73.71          &     68.21           &    57.84           
        \\  \hline
\multicolumn{2}{c|}{AsyCo original result:}                                                                   & \textbf{96.00} & \textbf{95.82} & \textbf{95.01} & \textbf{94.13} & \textbf{76.02} & \textbf{74.02} & \textbf{68.96} & \textbf{60.35} \\
\hline \bottomrule 
\end{tabular}}
\caption{Ablation study for the classification net $n_\theta$ and reference net $r_\phi$. }
\label{tab:ablation_all}
\end{table*}

Fig.~\ref{fig:cifar100_0.2} and~\cref{fig:cifar100_0.5} show the loss histograms after warmup for each subset in~\cref{tab:agreement_labels}. To compare with small-loss sample selection approaches, we adopt the sample-selection approach by DivideMix~\cite{li2020dividemix}  that is based on a Gaussian Mixture Model (GMM) to divide the training set into clean and noisy subsets (the vertical black dotted line is the threshold estimated by DivideMix).
These graphs show that the subsets' loss histograms are relatively consistent in different noise rates. 
Specifically, $\mathrm{C}$ always has the smallest loss values among all subsets, which shows that our multi-view sample selection is able to confidently extract clean samples. We also observe that $\mathrm{NY}$  has small loss values in both graphs. 
However, using $\mathrm{NY}$ as clean set does not produce promising performance, as shown in~\cref{tab:ablation_all}, row 'Small-loss subsets', which represents the use of almost all samples in C and NY as clean samples (since they are on the left-hand side of the GMM threshold).
This indicates that the small-loss samples in $\mathrm{NY}$ are likely to contain overfitted noisy-label samples, whereas our multi-view sample selection successfully avoids selecting these samples. In ~\cref{fig:acc_cifar100_0.2} and~\cref{fig:acc_cifar100_0.5}, we show the accuracy of the clean set selected by the GMM-based small-loss strategy of DivideMix and by our multi-view consensus during the training stages. We observe that multi-view selection performs consistently better than GMM in both graphs. 
We also validate the accuracy of the hidden clean label produced by the top ranked predictions of $r_{\phi}(.)$ by comparing the re-labelling produced by~\cref{eq:haty} versus no re-labelling (i.e., train $r_{\phi}(.)$ with the original training labels.)
Our multi-view re-labelling consistently improves the label accuracy overtime, which indicates the effectiveness of our method.

\cref{tab:ablation_all} shows a study on the selection of different subsets from~\cref{tab:agreement_labels} for the sample-selection when training the classification net $n_\theta(.)$.
First, we test the importance of classifying the samples in $\mathrm{RY}$ as clean for training $n_{\theta}(.)$ by, instead, treating these samples as noisy in Eq.~\eqref{eq:loss_training_net} (i.e., by setting $\mathbf{w}_i=0$). 
This new sample selection causes a large drop in performance for all cases, which suggests that $\mathrm{RY}$ contains informative samples that are helpful for training $n_{\theta}(.)$. 
Second, we test whether using the unmatched samples in $\mathrm{U}$ can improve model training, where we include them as clean or noisy samples by setting $\mathbf{w}_i=+1,0$, respectively. 
Both studies lead to worse results compared to the original AsyCo that discards $\mathrm{U}$ samples (see last row). 
Despite this result, we also notice that in low noise rates (0.2, 0.3), treating $\mathrm{U}$ as clean leads to slightly better accuracy than treating $\mathrm{U}$ as noisy. 
These results suggest that the high uncertainty and lack of view agreements by the samples in $\mathrm{U}$ lead to poor supervisory training signal, which means that discarding these samples is currently the best option. 
Finally, the histograms of~\cref{fig:ablation_loss} indicate that $\mathrm{NY}$ also contains small-loss samples. 
Therefore, we make the traditional small-loss assumption to train our AsyCo and use the subsets $\mathrm{C}$ and $\mathrm{NY}$ as clean and treat the other subsets as noisy. As shown in the "Small-loss subset" row of~\cref{tab:ablation_all}, the accuracy is substantially lower, which suggests that the small-loss samples may contain overfitted noisy-label samples.

We analyse the training of $r_\phi(.)$ with different training losses and re-labelling strategies in~\cref{tab:ablation_all}. 
We first study how the multi-label training loss provided by the BCE loss helps mitigate label noise
by training our reference net $r_{\theta}(.)$ with the CE loss  $\ell_{CE}(.)$ in Eq.~\eqref{eq:warmup_loss} and~\eqref{eq:training_reference_net}, while keeping the multi-view sample selection and re-labelling strategies unchanged. 
We observed that by training $r_{\theta}(.)$ with $\ell_{CE}(.)$ leads to a significant drop in accuracy for most cases, where for CIFAR10 with low noise rate (20\% and 30\%), $\ell_{CE}(.)$ maintains the accuracy of $\ell_{BCE}(.)$, but for larger noise rates, such as 40\% and 50\%, $\ell_{CE}(.)$ is not competitive with $\ell_{BCE}(.)$ because it reduces the prediction disagreements between $n_{\theta}(.)$ and $r_{\phi}(.)$, facilitating the overfitting to the same noisy-label samples by both models. 
For CIFAR100, $\ell_{CE}(.)$ leads to worse results than  $\ell_{BCE}(.)$ for all cases. 
These results suggest that to effectively co-teach two models with prediction disagreement, the use of different training strategies is an important component. 
Next, we study a training, where $r_{\phi}(.)$ is frozen after warmup, but we still train $n_{\theta}(.)$.
The result drops significantly which indicates that $r_{\phi}(.)$ needs to be trained in conjunction with  $n_{\theta}(.)$ to achieve reasonable performance. 
We study different re-labelling strategies by first setting $\hat{\mathbf{y}}_i=\tilde{\mathbf{y}}$ for training $r_{\phi}(.)$, which leads to comparable results for low noise rates, but worse results for high-noise rates, suggesting that that only training with $\tilde{\mathbf{y}}$ is not enough to achieve good performance. 
Finally, by setting $\hat{\mathbf{y}}_i=\mathbf{\tilde{y}}^{(n)}$, we notice better but slightly worse results than our proposed re-labelling from Eq.~\eqref{eq:haty}.

\section{Conclusion}
In this work, we introduced a new noisy label learning method called AsyCo. Unlike previous SOTA noisy label learning methods that train two models with the same strategy and select small-loss samples, AsyCo explores two different training strategies and use multi-view consensus for sample selection. We show in experiments that AsyCo outperforms previous methods in both synthetic and real-world benchmarks. In the ablation study, we explore various subset selection strategies for sample selection and re-labelling, which show the importance of our design decisions.  
For future work, we will explore lighter models for the reference net as only rank prediction is required. We will also explore out-of-distribution (OOD) samples in noisy label learning because our method currently assumes all samples are in-distribution. 

\label{sec:conclusion}

{\small
\bibliographystyle{ieee_fullname}
\bibliography{egbib}

\begin{thebibliography}{10}\itemsep=-1pt

\bibitem{arazo2019unsupervised}
Eric Arazo, Diego Ortego, Paul Albert, Noel O’Connor, and Kevin McGuinness.
\newblock Unsupervised label noise modeling and loss correction.
\newblock In {\em International conference on machine learning}, pages
  312--321. PMLR, 2019.

\bibitem{berthelot2019mixmatch}
David Berthelot, Nicholas Carlini, Ian Goodfellow, Nicolas Papernot, Avital
  Oliver, and Colin~A Raffel.
\newblock Mixmatch: A holistic approach to semi-supervised learning.
\newblock {\em Advances in neural information processing systems}, 32, 2019.

\bibitem{blum1998combining}
Avrim Blum and Tom Mitchell.
\newblock Combining labeled and unlabeled data with co-training.
\newblock In {\em Proceedings of the eleventh annual conference on
  Computational learning theory}, pages 92--100, 1998.

\bibitem{chen2021boosting}
Yingyi Chen, Xi Shen, Shell~Xu Hu, and Johan~AK Suykens.
\newblock Boosting co-teaching with compression regularization for label noise.
\newblock In {\em Proceedings of the IEEE/CVF Conference on Computer Vision and
  Pattern Recognition}, pages 2688--2692, 2021.

\bibitem{cheng2022instance}
De Cheng, Tongliang Liu, Yixiong Ning, Nannan Wang, Bo Han, Gang Niu, Xinbo
  Gao, and Masashi Sugiyama.
\newblock Instance-dependent label-noise learning with manifold-regularized
  transition matrix estimation.
\newblock In {\em Proceedings of the IEEE/CVF Conference on Computer Vision and
  Pattern Recognition}, pages 16630--16639, 2022.

\bibitem{deng2009imagenet}
Jia Deng, Wei Dong, Richard Socher, Li-Jia Li, Kai Li, and Li Fei-Fei.
\newblock Imagenet: A large-scale hierarchical image database.
\newblock In {\em 2009 IEEE conference on computer vision and pattern
  recognition}, pages 248--255. Ieee, 2009.

\bibitem{devlin2018bert}
Jacob Devlin, Ming-Wei Chang, Kenton Lee, and Kristina Toutanova.
\newblock Bert: Pre-training of deep bidirectional transformers for language
  understanding.
\newblock {\em arXiv preprint arXiv:1810.04805}, 2018.

\bibitem{garg2022instance}
Arpit Garg, Cuong Nguyen, Rafael Felix, Thanh-Toan Do, and Gustavo Carneiro.
\newblock Instance-dependent noisy label learning via graphical modelling.
\newblock {\em arXiv preprint arXiv:2209.00906}, 2022.

\bibitem{han2018co}
Bo Han, Quanming Yao, Xingrui Yu, Gang Niu, Miao Xu, Weihua Hu, Ivor Tsang, and
  Masashi Sugiyama.
\newblock Co-teaching: Robust training of deep neural networks with extremely
  noisy labels.
\newblock {\em Advances in neural information processing systems}, 31, 2018.

\bibitem{han2021trusted}
Zongbo Han, Changqing Zhang, Huazhu Fu, and Joey~Tianyi Zhou.
\newblock Trusted multi-view classification.
\newblock {\em arXiv preprint arXiv:2102.02051}, 2021.

\bibitem{he2015deep}
Kaiming He, Xiangyu Zhang, Shaoqing Ren, and Jian Sun.
\newblock Deep residual learningfor image recognition.
\newblock {\em ComputerScience}, 2015.

\bibitem{jiang2020beyond}
Lu Jiang, Di Huang, Mason Liu, and Weilong Yang.
\newblock Beyond synthetic noise: Deep learning on controlled noisy labels.
\newblock In {\em International Conference on Machine Learning}, pages
  4804--4815. PMLR, 2020.

\bibitem{jiang2018mentornet}
Lu Jiang, Zhengyuan Zhou, Thomas Leung, Li-Jia Li, and Li Fei-Fei.
\newblock Mentornet: Learning data-driven curriculum for very deep neural
  networks on corrupted labels.
\newblock In {\em International conference on machine learning}, pages
  2304--2313. PMLR, 2018.

\bibitem{krizhevsky2009learning}
Alex Krizhevsky, Geoffrey Hinton, et~al.
\newblock Learning multiple layers of features from tiny images.
\newblock 2009.

\bibitem{krizhevsky2017imagenet}
Alex Krizhevsky, Ilya Sutskever, and Geoffrey~E Hinton.
\newblock Imagenet classification with deep convolutional neural networks.
\newblock {\em Communications of the ACM}, 60(6):84--90, 2017.

\bibitem{li2020dividemix}
Junnan Li, Richard Socher, and Steven~CH Hoi.
\newblock Dividemix: Learning with noisy labels as semi-supervised learning.
\newblock {\em arXiv preprint arXiv:2002.07394}, 2020.

\bibitem{litjens2017survey}
Geert Litjens, Thijs Kooi, Babak~Ehteshami Bejnordi, Arnaud Arindra~Adiyoso
  Setio, Francesco Ciompi, Mohsen Ghafoorian, Jeroen~Awm Van Der~Laak, Bram
  Van~Ginneken, and Clara~I S{\'a}nchez.
\newblock A survey on deep learning in medical image analysis.
\newblock {\em Medical image analysis}, 42:60--88, 2017.

\bibitem{liu2020early}
Sheng Liu, Jonathan Niles-Weed, Narges Razavian, and Carlos Fernandez-Granda.
\newblock Early-learning regularization prevents memorization of noisy labels.
\newblock {\em Advances in neural information processing systems},
  33:20331--20342, 2020.

\bibitem{liu2015classification}
Tongliang Liu and Dacheng Tao.
\newblock Classification with noisy labels by importance reweighting.
\newblock {\em IEEE Transactions on pattern analysis and machine intelligence},
  38(3):447--461, 2015.

\bibitem{malach2017decoupling}
Eran Malach and Shai Shalev-Shwartz.
\newblock Decoupling" when to update" from" how to update".
\newblock {\em Advances in neural information processing systems}, 30, 2017.

\bibitem{paszke2019pytorch}
Adam Paszke, Sam Gross, Francisco Massa, Adam Lerer, James Bradbury, Gregory
  Chanan, Trevor Killeen, Zeming Lin, Natalia Gimelshein, Luca Antiga, et~al.
\newblock Pytorch: An imperative style, high-performance deep learning library.
\newblock {\em Advances in neural information processing systems}, 32, 2019.

\bibitem{patrini2017making}
Giorgio Patrini, Alessandro Rozza, Aditya Krishna~Menon, Richard Nock, and
  Lizhen Qu.
\newblock Making deep neural networks robust to label noise: A loss correction
  approach.
\newblock In {\em Proceedings of the IEEE conference on computer vision and
  pattern recognition}, pages 1944--1952, 2017.

\bibitem{ridnik2021asymmetric}
Tal Ridnik, Emanuel Ben-Baruch, Nadav Zamir, Asaf Noy, Itamar Friedman, Matan
  Protter, and Lihi Zelnik-Manor.
\newblock Asymmetric loss for multi-label classification.
\newblock In {\em Proceedings of the IEEE/CVF International Conference on
  Computer Vision}, pages 82--91, 2021.

\bibitem{shi2020multi}
Min Shi, Yufei Tang, Xingquan Zhu, and Jianxun Liu.
\newblock Multi-label graph convolutional network representation learning.
\newblock {\em IEEE Transactions on Big Data}, 2020.

\bibitem{simonyan2014very}
Karen Simonyan and Andrew Zisserman.
\newblock Very deep convolutional networks for large-scale image recognition.
\newblock {\em arXiv preprint arXiv:1409.1556}, 2014.

\bibitem{sindhwani2005co}
Vikas Sindhwani, Partha Niyogi, and Mikhail Belkin.
\newblock A co-regularization approach to semi-supervised learning with
  multiple views.
\newblock In {\em Proceedings of ICML workshop on learning with multiple
  views}, volume 2005, pages 74--79. Citeseer, 2005.

\bibitem{song2019selfie}
Hwanjun Song, Minseok Kim, and Jae-Gil Lee.
\newblock Selfie: Refurbishing unclean samples for robust deep learning.
\newblock In {\em International Conference on Machine Learning}, pages
  5907--5915. PMLR, 2019.

\bibitem{wang2017chestx}
Xiaosong Wang, Yifan Peng, Le Lu, Zhiyong Lu, Mohammadhadi Bagheri, and
  Ronald~M Summers.
\newblock Chestx-ray8: Hospital-scale chest x-ray database and benchmarks on
  weakly-supervised classification and localization of common thorax diseases.
\newblock In {\em Proceedings of the IEEE conference on computer vision and
  pattern recognition}, pages 2097--2106, 2017.

\bibitem{wei2020combating}
Hongxin Wei, Lei Feng, Xiangyu Chen, and Bo An.
\newblock Combating noisy labels by agreement: A joint training method with
  co-regularization.
\newblock In {\em Proceedings of the IEEE/CVF Conference on Computer Vision and
  Pattern Recognition}, pages 13726--13735, 2020.

\bibitem{xia2020part}
Xiaobo Xia, Tongliang Liu, Bo Han, Nannan Wang, Mingming Gong, Haifeng Liu,
  Gang Niu, Dacheng Tao, and Masashi Sugiyama.
\newblock Part-dependent label noise: Towards instance-dependent label noise.
\newblock {\em Advances in Neural Information Processing Systems},
  33:7597--7610, 2020.

\bibitem{xia2019anchor}
Xiaobo Xia, Tongliang Liu, Nannan Wang, Bo Han, Chen Gong, Gang Niu, and
  Masashi Sugiyama.
\newblock Are anchor points really indispensable in label-noise learning?
\newblock {\em Advances in Neural Information Processing Systems}, 32, 2019.

\bibitem{xiao2015learning}
Tong Xiao, Tian Xia, Yi Yang, Chang Huang, and Xiaogang Wang.
\newblock Learning from massive noisy labeled data for image classification.
\newblock In {\em Proceedings of the IEEE conference on computer vision and
  pattern recognition}, pages 2691--2699, 2015.

\bibitem{xu2021faster}
Youjiang Xu, Linchao Zhu, Lu Jiang, and Yi Yang.
\newblock Faster meta update strategy for noise-robust deep learning.
\newblock In {\em Proceedings of the IEEE/CVF Conference on Computer Vision and
  Pattern Recognition}, pages 144--153, 2021.

\bibitem{yao2021instance}
Yu Yao, Tongliang Liu, Mingming Gong, Bo Han, Gang Niu, and Kun Zhang.
\newblock Instance-dependent label-noise learning under a structural causal
  model.
\newblock {\em Advances in Neural Information Processing Systems},
  34:4409--4420, 2021.

\bibitem{young2018recent}
Tom Young, Devamanyu Hazarika, Soujanya Poria, and Erik Cambria.
\newblock Recent trends in deep learning based natural language processing.
\newblock {\em ieee Computational intelligenCe magazine}, 13(3):55--75, 2018.

\bibitem{yu2019does}
Xingrui Yu, Bo Han, Jiangchao Yao, Gang Niu, Ivor Tsang, and Masashi Sugiyama.
\newblock How does disagreement help generalization against label corruption?
\newblock In {\em International Conference on Machine Learning}, pages
  7164--7173. PMLR, 2019.

\bibitem{zhang2021understanding}
Chiyuan Zhang, Samy Bengio, Moritz Hardt, Benjamin Recht, and Oriol Vinyals.
\newblock Understanding deep learning (still) requires rethinking
  generalization.
\newblock {\em Communications of the ACM}, 64(3):107--115, 2021.

\bibitem{zhang2017mixup}
Hongyi Zhang, Moustapha Cisse, Yann~N Dauphin, and David Lopez-Paz.
\newblock mixup: Beyond empirical risk minimization.
\newblock {\em arXiv preprint arXiv:1710.09412}, 2017.

\bibitem{zhang2021learning}
Yikai Zhang, Songzhu Zheng, Pengxiang Wu, Mayank Goswami, and Chao Chen.
\newblock Learning with feature-dependent label noise: A progressive approach.
\newblock {\em arXiv preprint arXiv:2103.07756}, 2021.

\bibitem{zhu2021second}
Zhaowei Zhu, Tongliang Liu, and Yang Liu.
\newblock A second-order approach to learning with instance-dependent label
  noise.
\newblock In {\em Proceedings of the IEEE/CVF Conference on Computer Vision and
  Pattern Recognition}, pages 10113--10123, 2021.

\end{thebibliography}
}

\end{document}


\title{Supplementary Material \\ Asymmetric Joint Training with Multi-view Consensus \\for Noisy Label Learning}

\twocolumn[{
\maketitle
\begin{center}
    \captionsetup{type=figure}
    \centering
    
    \begin{subfigure}[b]{0.24\textwidth}
        \centering
        \includegraphics[width=\textwidth]{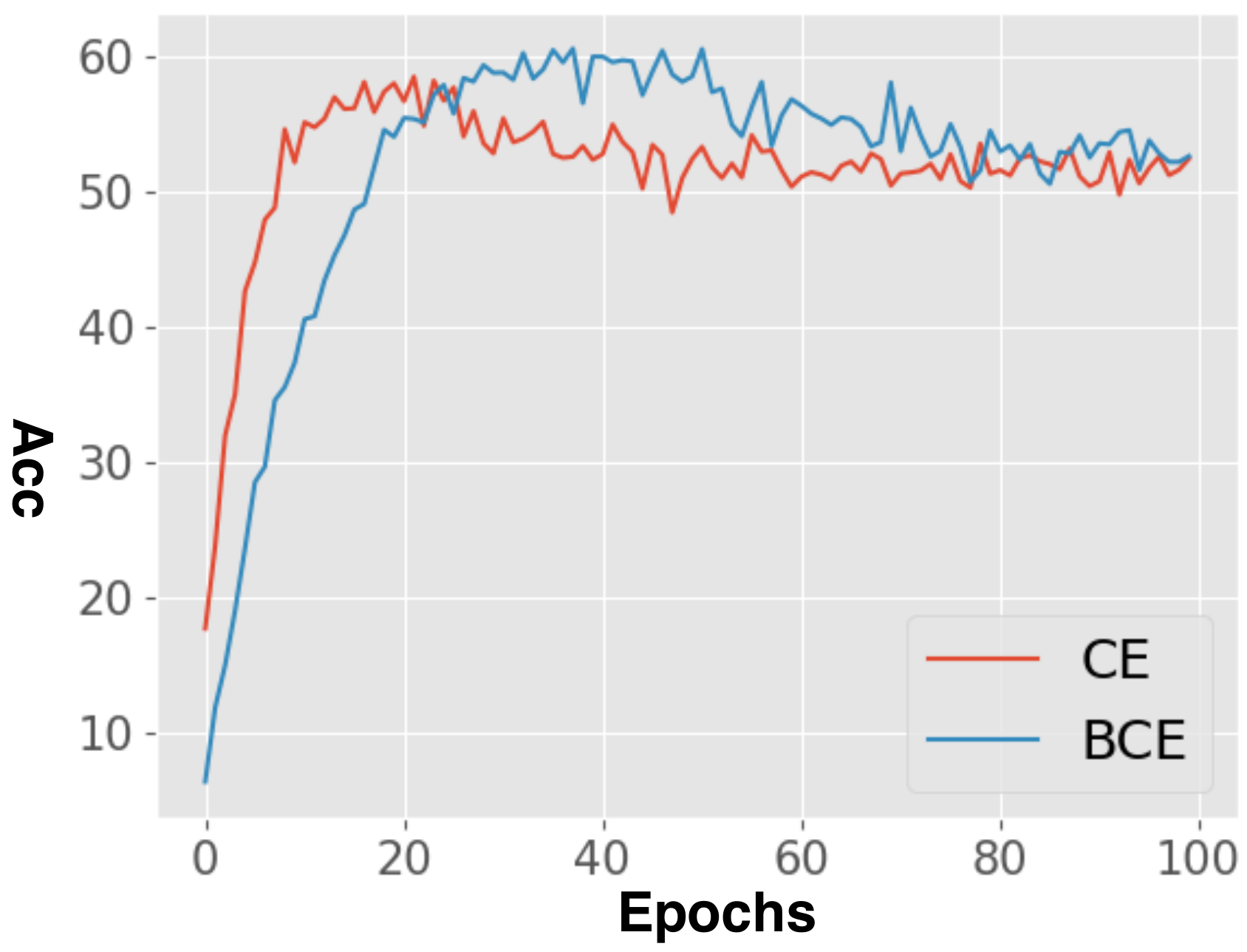}
        \caption[]%
        {CIFAR100 0.2 Accuracy} 
        \label{fig:cifar100_0.2}
    \end{subfigure}
    \hfill
    \begin{subfigure}[b]{0.24\textwidth}  
        \centering 
        \includegraphics[width=\textwidth]{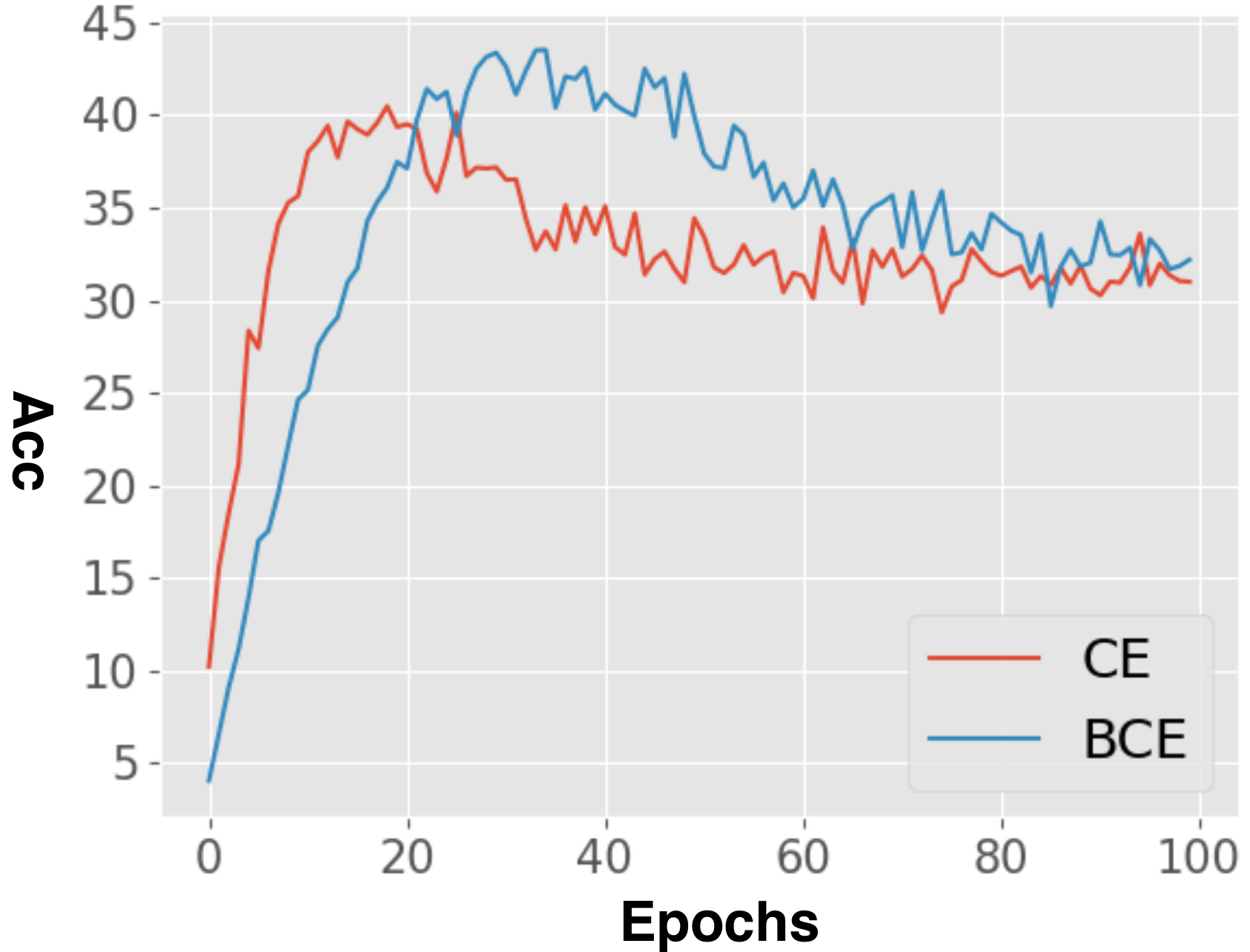}
        \caption[]%
        {CIFAR100 0.5 Accuracy}    
        \label{fig:acc_cifar100_0.2}
    \end{subfigure}
    \hfill
    \begin{subfigure}[b]{0.24\textwidth}   
        \centering 
        \includegraphics[width=\textwidth]{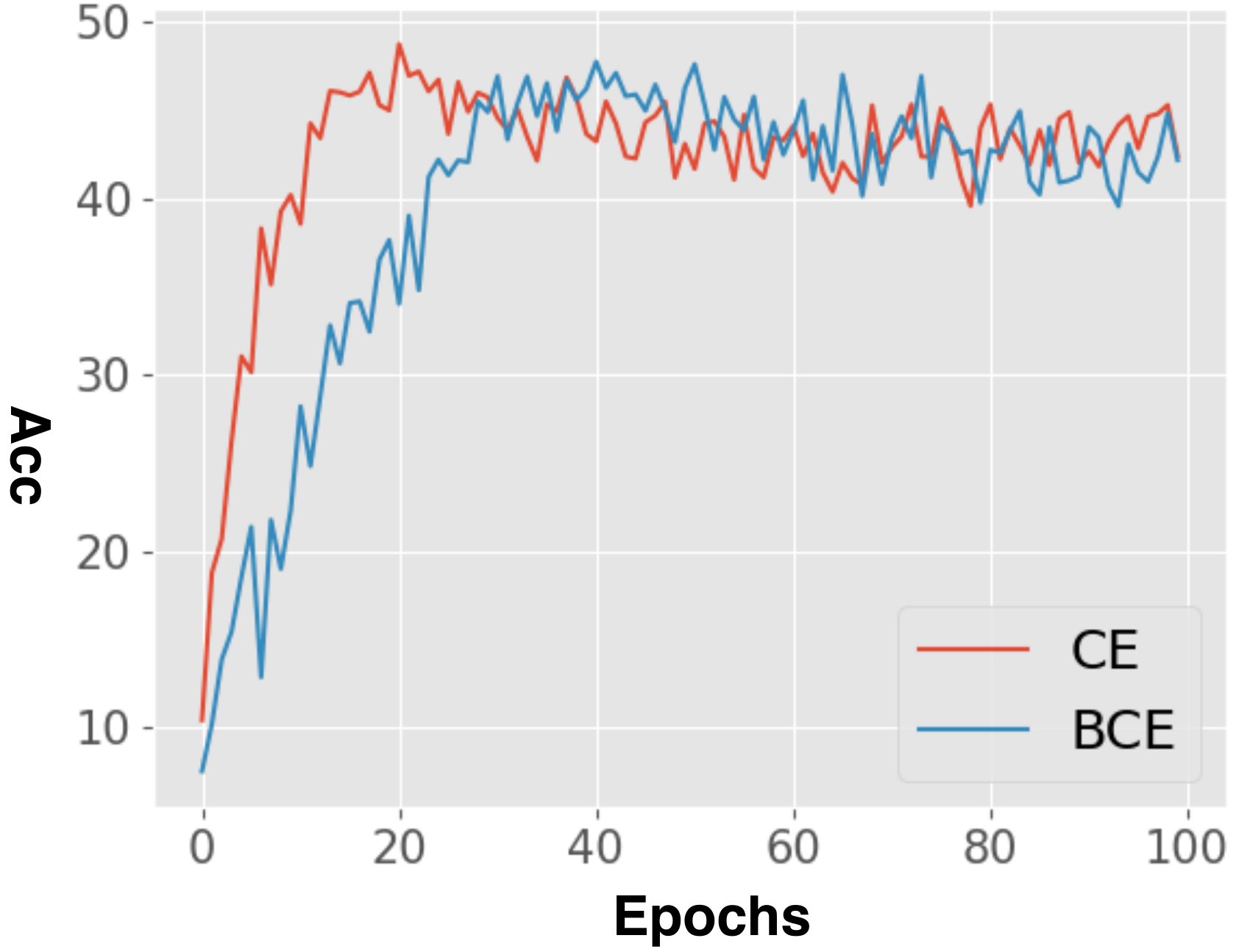}
        \caption[]%
         {Red Mini-ImageNet 0.2 Accuracy} 
         \label{fig:cifar100_0.5}
    \end{subfigure}
    \hfill
    \begin{subfigure}[b]{0.24\textwidth}
        \centering
        \includegraphics[width=\textwidth]{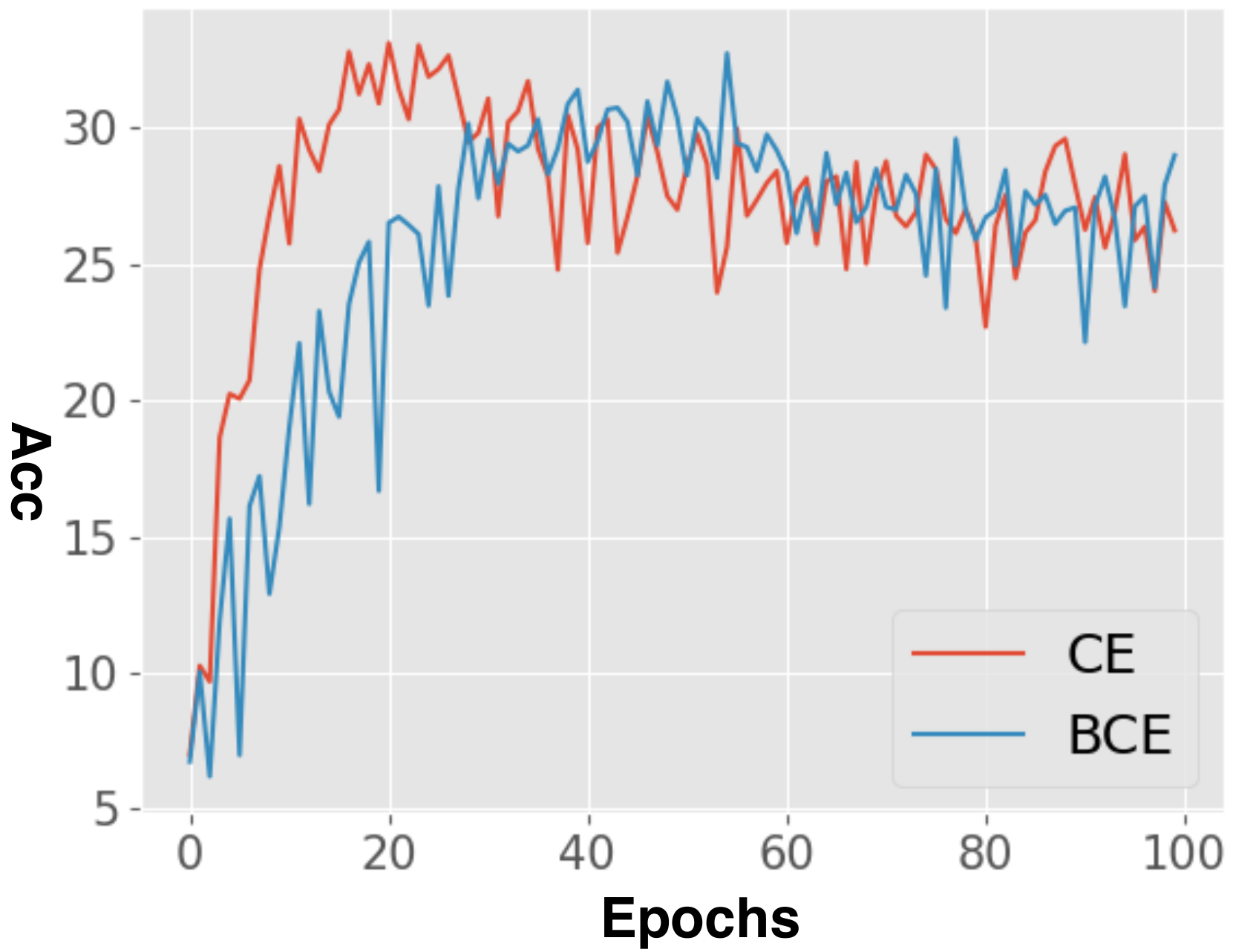}
        \caption[]%
         {Red Mini-ImageNet 0.8 Accuracy} 
        \label{fig:acc_cifar100_0.5}
    \end{subfigure}
    \caption{Comparison of the accuracy between a model trained with CE loss and another trained with BCE loss. The comparison is done for a training that lasts 100 epochs on CIFAR100 with 0.2/0.5 noise rates and Red Mini-ImageNet with 0.2/0.8 noise rates. 
    } 
    \label{fig:vis_ce_bce}
\end{center}
}]

\section{AsyCo Training Algorithm}

\cref{alg:asyjo} shows the main steps of our proposed AsyCo training algorithm.

\begin{algorithm}
  \caption{Asymmetric Joint Training Algorithm\label{alg:asyjo}}
  \begin{algorithmic}[1]
    \STATE \textbf{require: } Training net $n_\theta(.)$, reference net $r_\phi(.)$, training dataset $\mathcal{D}$ and number of training epochs $T$
    \STATE Warm-up $n_\theta(.)$ and $r_\phi(.)$ with Eq. 1 \\
    \WHILE{$t < T$}
    \STATE Compute $\tilde{\mathbf{y}}_i^{(n)}$ and $\tilde{\mathbf{y}}_i^{(r)}$ with Eq. 2 \\
    \STATE Categorize training samples from Tab. 1 \\
    \STATE Estimate sample selection latent variable $\mathbf{w}$, from Eq. 4, which classifies samples into clean or noisy\\
    \STATE Train $n_{\theta^{*}}(.)$ with Eq. 5 using $\mathbf{w}$ and $\mathcal{D}$ \\
    \STATE Estimate latent variable $\mathbf{\hat{y}}$, from Eq. 6, which re-labels the training samples with multiple labels  \\
    \STATE Train $r_{\phi^{*}}(.)$ with Eq. 7 using $\mathbf{\hat{y}}$ and $\mathcal{D}$
    \ENDWHILE
    \RETURN $n_{\theta^{*}}$
  \end{algorithmic}
\end{algorithm}

\section{Training Strategy Visualization}

In Fig.~\ref{fig:vis_ce_bce}, we show a visualisation of the testing accuracy differences for training two models, one with with CE loss and another with BCE loss, for 100 epochs on instance-dependent CIFAR100 with 0.2/0.5 noise rates and real-world Red Mini-ImageNet with 0.2/0.8 noise rates. We observe that CE and BCE show distinct training behaviours on both datasets and noise rates. Training with CE converges faster than BCE, but it also overfits more easily than BCE. Training with BCE takes longer to reach the same performance as CE, but it also overfits more slowly. This suggests that differences in training strategies can be explored for multi-view consensus selection.

\section{Sample Selection Time Comparison}
\begin{table}[t!]
\centering

\begin{tabular}{l|ccc}
\toprule \hline
Methods & GMM ~\cite{li2020dividemix}  & FINE~\cite{kim2021fine} & Ours \\ \hline
Time    & 17.2s & 34s  & 13s  \\ \hline \bottomrule
\end{tabular}
\caption{Time differences for each sample selection strategy on CIFAR10.}
\label{tab:time}
\end{table}

In Tab.~\ref{tab:time}, we show the running times of the small-loss based selection (GMM from DivideMix~\cite{li2020dividemix}), eigenvector-based selection (FINE~\cite{kim2021fine}), and our multi-view consensus selection. We observe that our approach is the most efficient, being 4s faster than small-loss and 20s faster than FINE. The reason is that our approach utilizes bit-wise AND operation for three views to partition the large training set into multiple subsets, which avoids multiple EM estimations. FINE performs the slowest because of the complexity involved in estimating the eigenvectors for each class.

{\small
\bibliographystyle{ieee_fullname}
\bibliography{egbib}
}